\documentclass[a4paper,fleqn]{cas-sc}

\usepackage[authoryear,longnamesfirst]{natbib}

\usepackage{amsmath}
\usepackage{amsthm}
\usepackage{algorithm}
\usepackage{algorithmicx}
\usepackage{algpseudocode}
\usepackage[title]{appendix}
\usepackage{textcomp}
\usepackage{tcolorbox}
\tcbuselibrary{breakable,skins}
\usepackage{cleveref}
\usepackage{chngcntr}
\usepackage{tabularx}
\usepackage{xspace}
\usepackage{booktabs}
\usepackage{float}
\usepackage{enumitem}
\usepackage{tikz}
\usetikzlibrary{arrows.meta,positioning}
\extrafloats{60}

\setlist{topsep=4pt,itemsep=2pt}

\newcommand{\fa}{Foam-Agent\xspace}
\newcommand{\fb}{CFDLLMBench\xspace}

\tcbset{breakable,enhanced}

\begin{document}

\let\WriteBookmarks\relax
\def\floatpagepagefraction{1}
\def\textpagefraction{.001}

\shorttitle{\fa: Automating Computational Fluid Dynamics Workflows}
\shortauthors{Yue et~al.}

\title[mode=title]{\fa: A Large Language Model-Based Multi-Agent Framework for Automating Computational Fluid Dynamics Workflows}

\author[1]{Ling Yue}
\credit{Conceptualization, Data curation, Formal analysis, Investigation, Methodology, Software, Validation, Writing -- original draft, Writing -- review and editing}
\author[2]{Nithin Somasekharan}
\credit{Conceptualization, Data curation, Formal analysis, Investigation, Methodology, Validation, Writing -- review and editing}
\author[1]{Tingwen Zhang}
\credit{Data curation, Methodology, Software, Validation, Writing -- review and editing}
\author[3]{Yadi Cao}
\credit{Conceptualization, Project administration, Supervision, Writing -- review and editing}
\author[4]{Zhangze Chen}
\credit{Validation, Visualization, Writing -- review and editing}
\author[5]{Shimin Di}
\credit{Conceptualization, Supervision, Writing -- review and editing}
\author[2]{Shaowu Pan}[orcid=0000-0002-2462-362X]
\cormark[1]
\ead{pans2@rpi.edu}
\credit{Conceptualization, Project administration, Funding acquisition, Resources, Supervision, Writing -- review and editing}


\affiliation[1]{organization={Department of Computer Science, Rensselaer Polytechnic Institute},
  city={Troy}, state={NY}, postcode={12180}, country={USA}}

\affiliation[2]{organization={Department of Mechanical, Aerospace, and Nuclear Engineering, Rensselaer Polytechnic Institute},
  city={Troy}, state={NY}, postcode={12180}, country={USA}}

\affiliation[3]{organization={Department of Computer Science and Engineering, University of California San Diego},
  city={La Jolla}, state={CA}, postcode={92093}, country={USA}}

\affiliation[4]{organization={College of Education, Zhejiang Normal University},
  city={Jinhua}, state={Zhejiang}, postcode={321004}, country={China}}

\affiliation[5]{organization={School of Computer Science and Engineering, Southeast University},
  city={Nanjing}, state={Jiangsu}, postcode={210096}, country={China}}

\cortext[1]{Corresponding author.}

\begin{abstract}
Computational fluid dynamics (CFD) has been the main workhorse of computational physics, yet its steep learning curve and fragmented, multi-stage workflow create significant barriers to entry. We present \fa, a multi-agent framework that leverages large language models (LLMs) to automate the end-to-end CFD workflow in OpenFOAM from a single natural-language prompt. \fa rests on three methodological contributions. First, a \emph{multi-index retrieval} scheme organizes domain knowledge along four complementary structural dimensions and selects indices by workflow stage, sharpening retrieval precision over conventional single-index retrieval-augmented generation. Second, \emph{dependency-aware file generation} is formulated as a topological traversal of the OpenFOAM case dependency graph, so that each configuration file is synthesized in the context of its already-generated predecessors, enforcing cross-file consistency. Third, a \emph{trajectory-conditioned reviewer loop} iteratively repairs failed runs by conditioning each correction on the accumulated error-and-diagnosis trajectory of its own previous attempts, applying at each step a minimal configuration edit that targets a reduced solver-error set. Around these contributions, six specialist agents span planning, meshing, file writing, execution, review, and visualization; \fa additionally exposes its capabilities through the Model Context Protocol as a deployment surface for external orchestrators. 
On FoamBench,
\fa achieves an 88.2\% execution success rate on the 110 Basic-tier tasks and 62.5\% on the out-of-distribution Advanced tier, $1.6\times$ and $5\times$ those of the prior MetaOpenFOAM framework, both with Claude~3.5~Sonnet, all without expert intervention, and we report a complementary field-level fidelity metric to distinguish crash-free execution from solution accuracy. These results demonstrate how the strategic harnessing of specialized multi-agent systems can reduce expertise barriers while preserving the rigor of solver-based simulation workflows.
\end{abstract}




\maketitle

\section{Introduction}

\begin{figure*}[hbpt]
\centering
\includegraphics[width=1.0\textwidth]{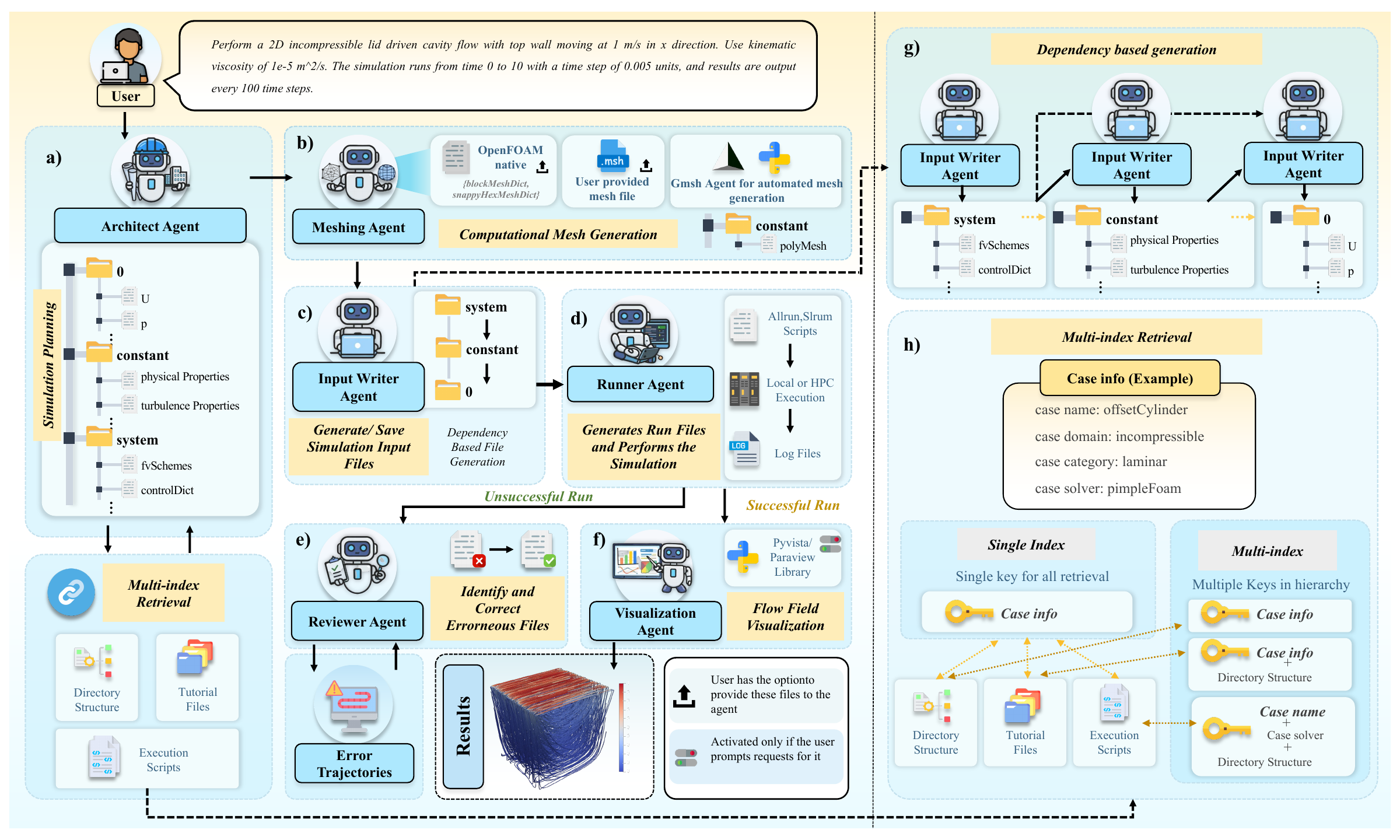}
\caption{\fa architecture. Starting from a natural-language request about CFD tasks, the framework plans the case structure, generates or imports the mesh, writes OpenFOAM files, executes the case, debugs failed runs, and produces post-processing visualizations. Panels (a)--(f) show the six agents; panel (g) shows dependency-aware file generation; panel (h) shows multi-index retrieval.}
\label{fig:overview}
\end{figure*}

Computational fluid dynamics (CFD)~\citep{Kumar2020ComputationalFD} underpins much of modern science and engineering. It is used to simulate fluid phenomena in applications that include aircraft~\citep{wang2014high, bravo2018conceptual}, wind-turbine design~\citep{aliyar2025directional, hartwanger20083d, lanzafame2013wind}, and blood flow in arteries~\citep{Kong2018SimulationOU, steinman2002image, zhang2014perspective}. It also lowers the cost of physical prototyping~\citep{Tu2007ComputationalFD}. 
However, mastering numerical simulations in OpenFOAM~\citep{jasak2007openfoam}, along with meshing tools such as Gmsh~\citep{geuzaine2009gmsh} and post-processing utilities, has a steep learning curve in practice. 
Such workflow requires several steps ranging from pre-processing,  such as geometry creation and meshing, context-dependent solver configuration, to post-processing for visualization~\citep{slotnick2014cfd}. 
As a result, this multi-stage process demands substantial domain expertise, and the resulting learning curve limits access for many researchers and engineers~\citep{Mari2014TheOT}.
Recent advances in generative AI, particularly Large Language Models (LLMs), have drawn our attention to the autonomous ``AI agents'' that can carry out multi-step scientific workflows~\citep{xi2025rise}. These agents can interpret high-level natural language commands, break down tasks, call software tools, and iteratively refine solutions~\citep{yao2022react}. Several domains have already adopted them for automation of scientific workflows. In chemistry, ChemCrow~\citep{m2024augmenting} automates synthesis planning by calling external chemical tools. In materials science, multi-agent systems have identified new high-performance alloys by reasoning over material databases~\citep{Ghafarollahi2025}. Across these cases, agents autonomously take over tedious repetitive setup work that previously required a domain specialist.
Within engineering, this trend has taken root most deeply in computational mechanics. MechGPT~\citep{Buehler2024MechGPT} introduced a domain-specialized LLM for mechanics knowledge, and MeLM~\citep{buehler2023melm} provided an early language-model formulation that solves forward and inverse mechanics problems. Multi-agent collaborations followed: MechAgents~\citep{ni2024mechagents} showed that teams of LLM agents can set up and solve mechanics problems and integrate knowledge across them, and ProtAgents~\citep{ghafarollahi2024protagents} coordinated LLM agents for knowledge retrieval, protein-structure analysis, and physics-based simulation to autonomously design and analyze de novo proteins with targeted mechanical properties. The same agentic approach has reached finite-element analysis (FEA), with frameworks like AutoFEA~\citep{hou2025autofea} and MooseAgent~\citep{zhang2025mooseagent} automating solver-setup workflows, and ALL-FEM~\citep{deotale2026allfem} utilizing a fine-tuned LLM for finite-element code generation. 
More broadly agentic systems have also appeared for further downstream tasks, such as model order reduction~\citep{guo2025cae}.

Within CFD specifically, LLM-based agents have begun to translate natural-language problem descriptions into executable OpenFOAM configuration files. Early systems such as MetaOpenFOAM~\citep{chen2025metaopenfoam} and OpenFOAMGPT~\citep{pandey2025openfoamgpt} show early promise on this task, typically by way of Retrieval-Augmented Generation (RAG)~\citep{lewis2020retrieval}: the model retrieves relevant examples from a knowledge base and produces plausible solver settings from existing case templates.
However, several limitations remain. First, their {workflow coverage is narrow}, focusing almost exclusively on isolated solver configuration while omitting critical pre- and post-processing stages. 
Second, they {generate configuration files independently}, failing to account for the rigid cross-file dependencies and structural hierarchy within an OpenFOAM case. 
Third, their {execution paradigm is strictly open-loop}, meaning they produce simulation files statically and cannot dynamically diagnose or correct runtime errors when a simulation fails.

In this work, we present \fa, a multi-agent framework for end-to-end CFD automation in OpenFOAM. First, where agentic designs range from decentralized systems~\citep{wang2026scienceclaw} that rediscover the workflow at run time to general-purpose coding agents that write solver inputs directly, \fa~instead binds each of six specialist agents to a canonical OpenFOAM stage: because the workflow is well understood and stable, encoding it explicitly lets each agent specialize rather than spend reasoning budget rediscovering the pipeline. Second, \fa~is built as domain-level \emph{harness engineering}, the design of the agent loop, tool interface, and context engineering around the model, so that its contributions are properties of the agent design rather than of any single transport layer and survive a change of retrieval backend or orchestration protocol. On this foundation we introduce three methodological advances for automated CFD workflow setup with LLM agents:
\begin{itemize}
\item \textbf{Multi-index retrieval} organizes OpenFOAM domain knowledge along four complementary structural dimensions (case metadata, directory structure, file contents, and execution scripts) and selects the index by current workflow stage. This improves retrieval precision over a flat single-index design.
\item \textbf{Dependency-aware file generation as a topological traversal} of the OpenFOAM case dependency graph: each file is generated in the verbatim context of its already-emitted predecessors, which enforces cross-file consistency and suppresses the hallucination of undefined symbols.
\item \textbf{Trajectory-conditioned reviewer loop}: an execution-grounded self-correction mechanism that diagnoses solver errors against the accumulated trajectory of its own previous attempts and applies a minimal configuration edit targeting a reduced error set on the next run.
\end{itemize}
These three advances are situated at the workflow layer: they concern how retrieval, file generation, and error correction are organized around the solver, rather than the numerical methods, discretization, or mesh technology inside it, which \fa~uses as provided by OpenFOAM. The remaining features of the framework, namely the Model Context Protocol (MCP)~\citep{anthropic2024mcp} interface, HPC execution support, external-mesh ingestion, and post-processing visualization, are engineering capabilities that broaden its practical use rather than methodological contributions. In this sense \fa~is complementary to recent agentic work in computational mechanics that operates at other layers, such as the LLM-empowered CAE agent for model order reduction~\citep{guo2025cae} and the fine-tuned finite-element system ALL-FEM~\citep{deotale2026allfem}, which advance the solver and model layers that the present workflow layer takes as given.

The remainder of this paper is organized as follows. \Cref{sec:method} describes the \fa~workflow, including the agent components, dependency-aware file generation, multi-index retrieval, and MCP interface. \Cref{sec:results} presents the experimental evaluation, covering overall performance, ablation studies, supported workflow capabilities, and case studies. \Cref{sec:conclusion} concludes the paper.

\section{Methods}\label{sec:method}

\subsection{Agent Components}
\label{sec:agents}

\begin{algorithm}
\caption{\fa~orchestration protocol; the Reviewer call is expanded in \Cref{alg:reviewer}.}
\label{alg:orchestration}
\begin{algorithmic}[1]
\Require Natural language requirement $R$, maximum iterations $M$
\Statex \textit{Symbols:} $\Pi=\{T_1,\dots,T_n\}$ is the file-generation plan produced by the Architect, where each task $T_i=(f_i,\rho_i,\sigma_i)$ specifies a target file path, a generation priority, and retrieved schematic constraints; $\mathcal{H}=[(\mathcal{E}_1,d_1),\dots,(\mathcal{E}_{t-1},d_{t-1})]$ is the append-only history of error sets and Reviewer diagnoses over correction iterations.
\Ensure Validated simulation configuration $\mathcal{S}^*$, Visualization $V$
\State $(\Pi, \mathcal{K}) \gets \operatorname{Architect}(R)$ \Comment{Plan files; retrieve reference $\mathcal{K}$}
\State $\mathcal{M} \gets \operatorname{MeshingAgent}(R)$ \Comment{Generate mesh (Native/Gmsh/External)}
\State $\mathcal{S}_0 \gets \operatorname{InputWriter}(\Pi, \mathcal{M}, \mathcal{K})$ \Comment{Generate initial case files}
\State $\mathcal{H} \gets [\,]$ \Comment{Initialize iteration history}
\For{$t \gets 1$ to $M$}
    \State $\mathcal{L}_t, \text{status} \gets \operatorname{Runner}(\mathcal{S}_{t-1})$ \Comment{Execute simulation and capture logs}
    \If{$\text{status} = \text{SUCCESS}$}
        \State $V \gets \operatorname{VisualizationAgent}(\mathcal{S}_{t-1}, R)$
        \State \Return $\mathcal{S}_{t-1}, V$
    \EndIf
    \State $\mathcal{E}_t \gets \Phi(\mathcal{L}_t)$ \Comment{Extract solver errors (\Cref{alg:runner})}
    \State $(\Delta \mathcal{S}_t, d_t) \gets \operatorname{Reviewer}(\mathcal{E}_t, \mathcal{S}_{t-1}, \mathcal{H}, \mathcal{K})$ \Comment{Trajectory-conditioned correction (\Cref{alg:reviewer})}
    \State $\mathcal{S}_t \gets \mathcal{S}_{t-1} \oplus \Delta \mathcal{S}_t$ \Comment{Apply configuration edit}
    \State $\mathcal{H} \gets \mathcal{H} \,\Vert\, [(\mathcal{E}_t, d_t)]$ \Comment{Append-only history}
\EndFor
\State \Return \text{FAILURE}
\end{algorithmic}
\end{algorithm}

\subsubsection{Architect Agent}
\label{sec:architect}
The Architect Agent translates a natural-language description into a structured, dependency-ordered file-generation plan through an explicit five-step pipeline, employing a RAG paradigm together with Pydantic~\citep{Colvin_Pydantic_Validation_2025} schemas for typed, validated intermediate representations. The pipeline runs in five steps: (i)~\emph{classification} of the prompt into a four-field Pydantic schema; (ii)~\emph{template selection} by FAISS retrieval over the multi-index (\Cref{sec:rag_retrieval}), a two-step query that returns the top-$k$ structurally matching tutorials and then re-queries the tutorial-details index for the full file contents of the top-ranked match; (iii)~\emph{file-manifest} decomposition of the retrieved tutorial into an explicit list of \texttt{(file\_name, folder\_name)} subtask pairs, augmented with case-specific files when the classification indicates them; (iv)~\emph{dependency layering} along the acyclic OpenFOAM \texttt{system}~$\succ$~\texttt{constant}~$\succ$~\texttt{0}~$\succ$~\texttt{Allrun} directory order, which every tutorial obeys; and (v)~\emph{emission}, in which the Input Writer (\Cref{sec:inputwriter}) generates the files in this layer order, each in the context of its already-emitted predecessors.

The classification schema has four categorical fields: \texttt{case\_name}, a free-form identifier; \texttt{case\_domain}, one of the 14 OpenFOAM tutorial-tree categories (\texttt{basic}, \texttt{combustion}, \texttt{compressible}, \texttt{discreteMethods}, \texttt{DNS}, \texttt{electromagnetics}, \texttt{financial}, \texttt{heatTransfer}, \texttt{incompressible}, \texttt{lagrangian}, \texttt{mesh}, \texttt{multiphase}, \texttt{resources}, \texttt{stressAnalysis}); \texttt{case\_category}, a solver-family identifier; and \texttt{case\_solver}, the specific solver name. The latter two fields are restricted at run time to values harvested from the tutorial corpus, and schema-level validation rejects malformed model output and re-prompts with a stricter directive. Turbulence class and time dependence (steady or transient) are not separate top-level fields; they are fixed implicitly by the tutorial retrieved in step~(ii).

Formally, the Architect maps the user-requirement space $\mathcal{R}$ to a structured task space $\Pi = \{T_1, T_2, \dots, T_n\}$, where $n$ is the number of files to be generated and each task $T_i$ is a tuple $(f_i, \rho_i, \sigma_i)$ with target file path $f_i$, generation priority $\rho_i$, and retrieved schematic constraints $\sigma_i$. This decomposition guarantees that subsequent generation respects the topological dependencies of the OpenFOAM case structure.

As a worked example, consider the request \emph{``simulate a laminar incompressible lid-driven cavity at $Re=100$''}. Step~(i) classifies it with \texttt{case\_name}~$=$~\texttt{lidDrivenCavity}, \texttt{case\_domain}~$=$~\texttt{incompressible}, and \texttt{case\_solver}~$=$~\texttt{icoFoam}. Step~(ii) retrieves the \texttt{cavity}/\texttt{icoFoam} tutorial as the top-1 match. Step~(iii) produces the manifest \{\texttt{controlDict}, \texttt{fvSchemes}, \texttt{fvSolution}, \texttt{transportProperties}, \texttt{blockMeshDict}, \texttt{0/U}, \texttt{0/p}, \texttt{Allrun}\}. Step~(iv) assigns these to four layers (layer~0~$=$~\{\texttt{controlDict}, \texttt{fvSchemes}, \texttt{fvSolution}, \texttt{blockMeshDict}\}, layer~1~$=$~\{\texttt{transportProperties}\}, layer~2~$=$~\{\texttt{0/U}, \texttt{0/p}\}, and layer~3~$=$~\{\texttt{Allrun}\}) and step~(v) emits them in that order.

\subsubsection{Meshing Agent}
\label{sec:meshing}
The Meshing Agent supports four mesh-generation pathways and autonomously selects one according to the user requirement.
\begin{itemize}
\item \texttt{blockMesh}. Structured, block-structured hexahedral meshes generated from a \texttt{blockMeshDict} that the agent writes; the agent then runs \texttt{blockMesh} to produce the \texttt{polyMesh} folder. This pathway is fully autonomous.
\item \texttt{snappyHexMesh}. Unstructured, body-fitted, hex-dominant meshes around STL geometries, generated from an agent-written \texttt{snappyHexMeshDict}; this pathway requires a watertight STL surface.
\item Gmsh~\citep{geuzaine2009gmsh}. Structured or unstructured meshes for complex geometries absent from standard tutorials. From a natural-language description of the physical domain and boundary names, the agent writes a Python script that builds the geometry and mesh through the Gmsh library, produces a \texttt{.msh} file, and converts it to OpenFOAM \texttt{polyMesh} format with \texttt{gmshToFoam}.
\item External \texttt{.msh}. A user-supplied, externally generated mesh is converted in place with \texttt{gmshToFoam}, after which the rest of the pipeline proceeds unchanged. The current external pathway is scoped to the Gmsh \texttt{.msh} format; additional external formats would require conversion-specific boundary-patch validation before being treated as fully supported.
\end{itemize}
Meshing for complex geometries is an open problem in CFD, and the Meshing Agent integrates existing tools rather than introducing a new meshing algorithm; it therefore inherits each tool's limits. The external-\texttt{.msh} pathway in particular is not a technical novelty but a deliberate human--AI design boundary: when the in-the-loop meshing tools fall short for a given geometry, the user can supply an expert mesh directly and the remaining agents continue without modification.

\subsubsection{Input Writer Agent}
\label{sec:inputwriter}
The Input Writer Agent realizes the dependency-aware file generation that is the second methodological contribution of this work. It emits the planned files as a topological traversal of the OpenFOAM case dependency graph $G=(V,E)$, where the vertices $V$ are configuration files and the edges $E$ are parameter dependencies. For example, files in the \texttt{0} directory depend on physical-property files in \texttt{constant}, which in turn depend on solver configuration in \texttt{system}. Because the OpenFOAM directory hierarchy fixes this layering (\texttt{system}~$\succ$~\texttt{constant}~$\succ$~\texttt{0}~$\succ$~\texttt{Allrun}), the traversal is realized by a stable four-tier priority sort over folder names rather than by re-deriving the graph at run time (\Cref{alg:inputwriter}).

\begin{algorithm}
\caption{Input Writer: fixed-layer topological emission}
\label{alg:inputwriter}
\begin{algorithmic}[1]
\Require Planned task list $\Pi$; retrieved tutorial reference $\mathcal{K}$
\Function{Priority}{$T_i$}
    \If{\textsc{Folder}($T_i$) $=$ \texttt{system}}
        \State \Return $0$
    \ElsIf{\textsc{Folder}($T_i$) $=$ \texttt{constant}}
        \State \Return $1$
    \ElsIf{\textsc{Folder}($T_i$) $=$ \texttt{0}}
        \State \Return $2$
    \Else
        \State \Return $3$
    \EndIf
\EndFunction
\State $\sigma \gets$ \textsc{StableSort}($\Pi$), keyed by \textsc{Priority} \Comment{topological order}
\State $\mathcal{C} \gets [\,]$ \Comment{accumulator of emitted files}
\For{$T_i \in \sigma$}
    \State $C_i \gets \operatorname{LLM}_{\mathrm{Writer}}(T_i,\ \mathcal{K},\ \mathcal{C})$ \Comment{$\mathcal{C}$: emitted predecessors}
    \State $\mathcal{C} \gets \mathcal{C} \,\Vert\, [C_i]$
\EndFor
\State $A \gets \operatorname{LLM}_{\mathrm{Allrun}}(\mathcal{C},\ \mathcal{K},\ \textsc{CommandHelp})$ \Comment{second LLM call: Allrun script}
\State \Return $\mathcal{C} \,\Vert\, [A]$
\end{algorithmic}
\end{algorithm}

For the $i$-th file the Architect supplies a task specification $T_i=(f_i,\rho_i,\sigma_i)$, and the file content $C_i$ is generated as
\begin{equation}
\label{eq:depgen}
C_i = \operatorname{LLM}\!\left(T_i,\ \mathcal{K},\ \{C_j : j \prec i\}\right),
\end{equation}
where $j \prec i$ denotes the files emitted before $i$ in topological order. More generally, the prompt assembled for each LLM call is the concatenation of four distinct components (the complete agent prompts are listed in \Cref{app:a}),
\begin{equation}
\label{eq:prompt}
\mathrm{prompt}_i = T_i \,\oplus\, \mathcal{K} \,\oplus\, \{C_j : j \prec i\} \,\oplus\, \mathcal{H},
\end{equation}
in which $T_i$ is the file task specification; $\mathcal{K}$ is the \emph{retrieved knowledge}, namely the top-ranked tutorial reference returned by the multi-index retrieval module of \Cref{sec:rag_retrieval} (shared across all Input Writer calls for a case); $\{C_j : j \prec i\}$ is the verbatim content of the already-emitted predecessor files; and $\mathcal{H}$ is the iteration history, which is empty during the initial Input Writer pass and grows only over Reviewer iterations (\Cref{sec:reviewer}). Injecting the predecessor files \emph{verbatim}, rather than as paraphrased summaries, anchors patch names, the solver name, and transport-property entries to the exact upstream tokens, which enforces parameter continuity and suppresses the hallucination of undefined symbols. Once all case files are emitted, a second LLM call generates the \texttt{Allrun} execution script from the accumulated files and the command documentation. Because the agents operate on configuration files rather than mesh data, the per-call input context remains small in practice, averaging $9$--$19$k tokens on the benchmark (\Cref{sec:cost}), so context-window overflow does not arise and no context compaction is required.

\subsubsection{Runner Agent}
\label{sec:runner}
The Runner Agent interfaces with the OpenFOAM execution environment: it prepares the simulation (cleaning artifacts, setting up output capture) and executes the \texttt{Allrun} script either on the user's \emph{local machine} or on an \emph{HPC cluster} (\Cref{fig:cavity_hpc}), as specified in the request. For HPC runs the agent generates a Slurm script, submits the job with the provided account, and monitors it to completion~\citep{openfoam_hpc_tc2025}; partition, node count, and processes-per-node are taken from the prompt or inferred from the mesh decomposition.

At the end of a run the Runner extracts solver errors for the Reviewer. This error detection is the pattern-matching function $\Phi: L \rightarrow \mathcal{E}$ that maps the execution logs $L$ to a set of structured error records $\mathcal{E}$. $\Phi$ is \emph{not} backed by a preloaded error catalog: it is a single-regex extractor that captures OpenFOAM's canonical fatal-error prefix from each solver log file, leaving the \emph{classification} of the captured fragment to the LLM Reviewer (\Cref{alg:runner}).

\begin{algorithm}
\caption{Runner: canonical-prefix error extraction}
\label{alg:runner}
\begin{algorithmic}[1]
\Require Case directory $D$; canonical regex \texttt{r = re.compile(r"ERROR:(.*)", re.DOTALL)}
\State $\mathcal{E} \gets \emptyset$
\For{each log file $\ell$ in $D$ (file name with prefix \texttt{log})}
    \State $c \gets \textsc{Read}(\ell)$
    \If{\texttt{r} matches $c$}
        \State $e \gets$ matched fragment after \texttt{ERROR:}
        \State $\mathcal{E} \gets \mathcal{E} \cup \{(\,\ell,\ e\,)\}$
    \EndIf
\EndFor
\State \Return $\mathcal{E}$
\end{algorithmic}
\end{algorithm}

\paragraph{Coverage.} OpenFOAM standard solvers abort through the \texttt{FatalErrorIn} and \texttt{FatalIOErrorIn} macros, which emit a line containing the canonical \texttt{ERROR:} substring into the solver log. This single substring captures the diagnosable evidence for the dominant failure classes: dictionary syntax errors, missing-file errors, boundary-condition mismatches, and thermophysical-bound aborts. Two classes systematically bypass the prefix: runner-script-level failures that abort before solver invocation (e.g.\ \texttt{mpirun} option errors), and bare floating-point exceptions whose stack trace contains no \texttt{ERROR:} line.

\paragraph{Update mechanism.} Because there is no catalog, there is nothing to maintain: any new OpenFOAM error type is supported automatically as long as the solver emits the canonical \texttt{ERROR:} prefix, after which the LLM Reviewer classifies and repairs the fragment with no agent-side code change. Keeping the runner-side surface deliberately narrow lets the LLM absorb the open vocabulary of solver-specific error wordings.

\paragraph{Failure modes.} Errors that bypass the \texttt{FatalErrorIn} macros (\texttt{mpirun} option errors, \texttt{decomposePar} configuration failures, and bare floating-point-exception traces) fall outside this extractor and are not diagnosable from the solver log alone. When a log contains several \texttt{ERROR:} fragments, the \texttt{re.DOTALL} flag returns one contiguous match per log (the tail starting at the first occurrence), which the Reviewer then decomposes internally during diagnosis.

\subsubsection{Reviewer Agent}
\label{sec:reviewer}
The Reviewer Agent drives the execution-grounded correction loop that is the third methodological contribution of this work. Within each orchestration iteration, indexed by the iteration counter $t$, when a run fails, the Reviewer node is invoked as three sequential LLM calls (\Cref{alg:reviewer}): an \emph{analysis} call produces a diagnosis $d_t$ from the extracted error set $\mathcal{E}_t$; a \emph{plan} call turns the diagnosis into a minimal rewrite plan $P_t$, the smallest set of file changes that targets a reduced error set on the next run; and a \emph{rewrite} call realizes the plan into the configuration edit $\Delta\mathcal{S}_t$ (a set of file changes, not an OpenFOAM boundary patch). The orchestration loop (\Cref{alg:orchestration}) then re-executes the case and repeats until the error set is empty or the iteration cap $M$ is reached.

\begin{algorithm}
\caption{Reviewer node: trajectory-conditioned correction step}
\label{alg:reviewer}
\begin{algorithmic}[1]
\Require Error set $\mathcal{E}_t$; case state $\mathcal{S}_{t-1}$; history $\mathcal{H}$; retrieved reference $\mathcal{K}$
\Ensure Configuration edit $\Delta\mathcal{S}_t$; diagnosis $d_t$
\State $d_t \gets \operatorname{LLM}_{\mathrm{Analyze}}(\mathcal{E}_t,\ \mathcal{S}_{t-1},\ \mathcal{K},\ \mathcal{H})$ \Comment{diagnose the error set}
\State $P_t \gets \operatorname{LLM}_{\mathrm{Plan}}(\mathcal{E}_t,\ \mathcal{S}_{t-1},\ d_t)$ \Comment{minimal rewrite plan: smallest file-change set}
\State $\Delta\mathcal{S}_t \gets \operatorname{LLM}_{\mathrm{Rewrite}}(\mathcal{S}_{t-1},\ P_t)$ \Comment{realize the plan into corrected files}
\State \Return $(\Delta\mathcal{S}_t,\ d_t)$
\end{algorithmic}
\end{algorithm}

We describe this loop as a trajectory-conditioned reviewer loop: across iterations it conditions each correction on the accumulated trajectory of its own previous attempts, the error logs and diagnoses recorded in the history $\mathcal{H}$ (the \emph{Iteration history} paragraph below), so that it learns from and avoids its earlier failed strategies, an execution-grounded form of self-reflection~\citep{shinn2023reflexion}. Within a single iteration the correction itself is kept minimal: the plan call is instructed to propose the smallest rewrite set, and the rewrite call to return only files that require modification or addition, so the per-step configuration edit $\Delta\mathcal{S}_t$ (a set of file changes, not an OpenFOAM boundary patch) targets a reduced error set with the fewest changes. The per-step edit is chosen greedily with respect to the current error set $\mathcal{E}_t$, while the diagnosis that produces it is conditioned on the full trajectory $\mathcal{H}$; the error reduction it targets is an \emph{empirical} target verified by re-execution rather than a guaranteed monotone descent, because an OpenFOAM error can be masked by an earlier failure and surface only after an edit is applied. The term ``optimization'' here refers to this minimal-edit objective and re-execution target, not to a numerical solver with an analytic convergence guarantee.

\paragraph{Corrective-action repertoire.} The Reviewer is not restricted to time-step reduction; it can issue five categories of corrective action: \emph{(i)}~boundary-condition correction, changing a patch type or value (for example, replacing a plain \texttt{wall} with a constraint-type wall when a buoyant solver requires it); \emph{(ii)}~discretization-scheme swaps in \texttt{fvSchemes} (for example, \texttt{Gauss linear}~$\to$~\texttt{Gauss limitedLinear} for shock-dominated cases); \emph{(iii)}~time-step and Courant-number adjustment via \texttt{controlDict}; \emph{(iv)}~mesh regeneration, re-invoking the Meshing Agent (including refinement-level changes for \texttt{blockMesh} and \texttt{snappyHexMesh}) when \texttt{checkMesh} flags excessive skewness, non-orthogonality, or aspect ratio; and \emph{(v)}~solver or turbulence-model replacement (for example, \texttt{pisoFoam}~$\to$~\texttt{simpleFoam} for a steady case). One scope clarification follows: the current loop can trigger mesh regeneration and mesh-parameter adjustment, but it does not modify the CAD or geometry definition itself in response to a downstream solver failure; closed-loop geometry modification is left as a complex-geometry extension.

\paragraph{Iteration history.} To diagnose against earlier attempts, the Reviewer maintains an append-only history $\mathcal{H}$. At each iteration $t$, an \texttt{<Error\_Logs>} block holding $\mathcal{E}_t$ and a \texttt{<Review\_Analysis>} block holding the diagnosis $d_t$ are appended, both wrapped in an \texttt{<Attempt>} element. On each Reviewer call the accumulated list is concatenated and wrapped in a \texttt{<history>} block alongside the current error logs, the current case files, and the retrieved tutorial reference. The history is kept \emph{flat and append-only}, rather than digested into a summary, because Reviewer trajectories frequently close non-monotonically: a single fix can expose a previously masked downstream error, so the outstanding error set can grow at intermediate steps even when the run eventually succeeds. In such trajectories the Reviewer needs the full record of which diagnoses were already tried, so it does not re-apply a failed strategy, and the full error-log trace, so it can recognize a re-emerging error pattern, exactly the evidence that a digest would compress away.

\subsubsection{Visualization Agent}
\label{sec:visualization}
After the Runner produces a successful run, \fa~checks whether the prompt requested a visualization. If so, the Visualization Agent identifies the target field variables and generates a Python script (using either the PyVista~\citep{sullivan2019pyvista} library or ParaView Python routines~\citep{ayachit2015paraview}, at the user's choice) to render them (\Cref{fig:overview}). It then executes the script and, like the Reviewer, repairs any errors until the visualizations are saved as PNG files in the run directory, up to a user-configurable retry limit.

\subsection{Multi-Index Retrieval}
\label{sec:rag_retrieval}

\begin{algorithm}
\caption{Multi-index retrieval with deterministic reranking}
\label{alg:rag}
\begin{algorithmic}[1]
\Require Query $q$ with classification tuple; workflow stage $s$; recall depth $k$ (default $3$)
\Ensure Retrieved tutorial reference $\mathcal{K}$
\State $\mathcal{I}_s \gets \textsc{SelectIndex}(s)$ \Comment{stage-specific FAISS index}
\State $\mathcal{D} \gets \textsc{TopK}\big(\mathcal{I}_s,\ \textsc{Embed}(q),\ k\big)$ \Comment{top-$k$ recall; no similarity cutoff}
\State sort $\mathcal{D}$ ascending by the lexicographic key
\Statex \quad $\big(-[\textsc{domain match}],\ -[\textsc{solver match}],\ -\textsc{cosine}\big)$
\State $\mathcal{K} \gets$ file contents of the top-1 candidate of sorted $\mathcal{D}$
\State \Return $\mathcal{K}$
\end{algorithmic}
\end{algorithm}

The first methodological contribution of \fa~is a multi-index retrieval system that segments OpenFOAM domain knowledge into specialized indices, each optimized for a specific phase of the simulation workflow. Compared with a conventional flat single-index RAG system, this structural decomposition sharpens retrieval precision; the gain is quantified directly by the ablation of \Cref{sec:ablation}.

\paragraph{Retrieval algorithm.} An earlier prototype of this module filtered candidates with a cosine-similarity cutoff $\tau$. Ahead of the released implementation we replaced it with the calibration-free procedure of \Cref{alg:rag}, because a fixed $\tau$ required per-solver tuning that scaled poorly across the heterogeneous OpenFOAM case-family distribution: it over-filtered on rare solver families (returning no candidate) and under-filtered on common ones (returning structurally distant candidates). The current algorithm exposes only two semantically meaningful knobs, both pinned in code: a top-$k$ recall depth ($k=3$ by default) and a deterministic lexicographic rerank. Given the Architect's four-field classification, FAISS returns the top-$k$ candidates from the stage-selected index; these are reranked by the key $\big(-[\texttt{case\_domain}~\text{match}],\ -[\texttt{case\_solver}~\text{match}],\ -\text{cosine}\big)$, where $[\cdot]$ is the Iverson bracket (one if the enclosed predicate holds, zero otherwise). The cosine score therefore acts only as a tiebreaker after the exact-match flags on \texttt{case\_domain} and \texttt{case\_solver}, and is never compared against a numeric cutoff. The top-1 reranked candidate is selected unconditionally and its file contents become the retrieved reference $\mathcal{K}$. The operative sensitivity question is thus the structural decomposition itself (multi-index versus flat single index), studied in \Cref{sec:ablation}; a sweep of the discrete recall depth $k\in\{1,3,5\}$ is left as a follow-up experiment.

\subsubsection{Knowledge Base Construction}
\label{sec:kb}
The knowledge base is built from the OpenFOAM tutorial corpus through a five-stage pipeline: (i)~\emph{source} selection of the tutorial corpus; (ii)~\emph{structural parsing} of each case into typed dictionary triples; (iii)~\emph{per-dimension extraction} across four complementary structural dimensions; (iv)~\emph{category tagging} of each tutorial; and (v)~\emph{indexing} into four FAISS indices. The source is all OpenFOAM Foundation v10 tutorials ($\approx 200$ cases) under \texttt{\$WM\_PROJECT\_DIR/tutorials}, spanning the major solver families (incompressible, compressible, multiphase, reactive, heat transfer, magneto-hydrodynamic), turbulence regimes (laminar, RAS, LES), and mesh-generation pathways. A lightweight parser walks each case directory and extracts \texttt{(key, value, type)} triples from the OpenFOAM dictionary files via top-level brace matching; it is targeted rather than a full grammar over the OpenFOAM dictionary language, capturing exactly the patterns the four retrieval dimensions require. The four dimensions are \emph{case metadata} (the solver name from \texttt{controlDict}, physics tags from the \texttt{Allrun} header, and the turbulence model from \texttt{constant/}\allowbreak\texttt{momentumTransport}), \emph{directory structure} (a bag-of-paths over the case file tree), \emph{file contents} (full file text together with the parsed triples), and \emph{execution scripts} (the \texttt{Allrun} command sequence). The four dimensions are correlated rather than independent, since a case's directory structure, file contents, and execution script all reflect its solver family. They are complementary in the sense that each serves a different retrieval stage: the design goal is this separation of concerns, so that each stage queries the index whose granularity matches its query, rather than a decomposition into statistically independent factors. Each tutorial is then annotated with the four categorical fields \texttt{case\_name}, \texttt{case\_domain}, \texttt{case\_category}, and \texttt{case\_solver}; the union of observed values is aggregated into a statistics file that constrains the Architect's run-time classification to values that actually occur in the tutorial tree, in an LLM-assisted pass with expert review. Finally, the four dimensions populate four FAISS~\citep{douze2025faiss} indices, one per dimension, queried at recall depth $k=3$; embeddings are computed locally with the Qwen3-Embedding-0.6B model, which removes the external embedding-API dependency without altering the indexing architecture.
The four indices serve distinct roles: the Tutorial Structure Index identifies structural templates; the Tutorial Details Index supplies boundary conditions, numerical schemes, and physical models; the Execution Scripts Index supplies command sequences; and the Command Documentation Index supplies utility usage and parameter selection.

\subsection{The Model Context Protocol Deployment Surface}
\label{sec:mcp}

Beyond the three methodological contributions above, \fa~exposes all of its agent functions through the Model Context Protocol~\citep{anthropic2024mcp} so that external orchestrators can consume it as a composable component~\citep{ouyang2025code2mcp}. We are explicit that MCP is a \emph{deployment surface} (an ecosystem-participation choice) rather than a research contribution: nothing inside \fa~depends on MCP, and the three contributions of \Cref{sec:method} hold independently of it.

What MCP provides is interface unification. Standard function calling could expose the same agent functions, but reaching each orchestrator would then require a bespoke adapter; MCP collapses this to a single server, published once and consumed by any MCP-capable client. For a single, fixed orchestrator, MCP offers little beyond ordinary function calling; its value emerges across orchestrators, where one self-describing server replaces the separate adapter that each additional client would otherwise require. In our deployment the same unmodified server is driven by both Cursor and Claude Code. This is the interoperability rationale that standardized formats already serve elsewhere in computational science, where CGNS~\citep{cgns} standardizes the exchange of mesh and solution data and preCICE~\citep{precice2016} standardizes black-box solver coupling, so that independently developed tools interoperate without pairwise adapters.

The function surface (atomic, stateful, workflow-decoupled functions identified by \texttt{case\_id} and \texttt{job\_id}) is detailed in \Cref{app:d}. When the orchestrator itself drives the call sequence, we implement it as a stateful LangGraph~\citep{langgraphdocs} graph whose nodes are MCP calls and whose edges are conditional transitions, with Pydantic~\citep{Colvin_Pydantic_Validation_2025} schemas enforcing typed, validated I/O at every boundary.


\paragraph{\fa~in the agentic-AI ecosystem.} \fa~is designed to be a plug-and-play CFD component for higher-level scientific-discovery ecosystems such as ScienceClaw~\citep{wang2026scienceclaw}. In our deployment a \emph{Skill} is layered on top of the MCP server: the Skill is a prompt-level wrapper that supplies an orchestrator with task context and usage conventions, while the underlying tool surface remains the MCP-exposed agent functions. The two layers serve different purposes (MCP is the tool-exposure protocol, the Skill is a higher-level prompt-plus-tool composition), and either can be consumed by orchestrators such as Cursor, Claude Code, or ScienceClaw. One consequence is important: when \fa~is consumed as a component, the orchestrator inherits the underlying capabilities (planning, meshing, file writing, execution, review, visualization) but \emph{not} the workflow design itself: planning and the iterative correction loop migrate from \fa~into the external orchestrator.

In that setting the iteration policy is owned by the orchestrator rather than by \fa: a strong orchestrator repeatedly invokes the repair function until the case completes, whereas a weaker one may exit prematurely, and standalone \fa~retains a deterministic loop up to its configured iteration cap. A measured MCP-versus-no-MCP performance ablation would be uninformative by construction: the MCP wrapper invokes the same in-process agent functions with the same arguments, so the OpenFOAM trajectory and the metrics of \Cref{sec:results} are identical with or without it; the substantive difference is integration cost, namely the number of orchestrator-specific adapters, not solution quality. Protocol-level details, namely per-call latency and the failure modes at the protocol boundary, are deferred to \Cref{app:d}.

\section{Results}\label{sec:results}

We evaluate \fa~on FoamBench, the end-to-end OpenFOAM simulation benchmark of the \fb~suite~\citep{somasekharan2025cfdllmbench}. FoamBench is organized into a \emph{Basic} tier of 110 cases across 11 physics scenarios and a harder \emph{Advanced} tier of 16 hand-crafted cases. Each case is specified by a natural-language prompt giving the problem description, physical scenario, geometry, solver, boundary conditions, and simulation parameters.

\paragraph{Evaluation metrics.} Because crash-free execution and physical correctness are distinct properties, we report two precise metrics throughout, following the \fb~convention~\citep{somasekharan2025cfdllmbench}.
\begin{itemize}
\item $M_{\mathrm{exec}}\in\{0,1\}$ is the \emph{execution-success} metric: it is $1$ if and only if the OpenFOAM solver returns exit code $0$, writes a final time step, and produces no NaN or Inf in the residuals (the ``End'' marker in the solver log).
\item $M_{\mathrm{NMSE}}\in\{0,0.5,1\}$ is a \emph{bucketed field-level fidelity} metric: the per-case normalized mean-squared error (NMSE) against the ground-truth solution is scored $1$ if $\mathrm{NMSE}\le 0.1$, $0.5$ if $0.1<\mathrm{NMSE}\le 0.3$, and $0$ otherwise. NMSE is the squared $L_2$ norm of the field deviation normalized by the squared $L_2$ norm of the ground-truth field, computed field-wise ($U$, $p$, $\rho$, $T$ where present) and averaged.
\end{itemize}
We define \emph{complete success} as $M_{\mathrm{exec}}\wedge(M_{\mathrm{NMSE}}{=}1)$. Byte-level identity to a human-expert reference is explicitly \emph{not} the success criterion: two OpenFOAM configurations can be syntactically distinct yet numerically equivalent in the regime of interest (for example, \texttt{fixedValue} versus \texttt{codedFixedValue} for a constant boundary condition, or two distinct stable time steps), and the NMSE-based metric correctly credits these as equivalent. Every headline $M_{\mathrm{exec}}$ figure below is co-reported with its $M_{\mathrm{NMSE}}$.

\subsection{Baselines and Evaluation Setup}
\label{sec:baselines}
We compare \fa~against MetaOpenFOAM~\citep{chen2025metaopenfoam}, a representative multi-agent framework for automating OpenFOAM workflows. To separate the contribution of agentic orchestration from raw LLM capability, we additionally include a \emph{zero-shot LLM} baseline (a single LLM call with no agents, retrieval, or Reviewer loop) as a framework-free lower bound. Both frameworks are evaluated with Claude 3.5 Sonnet~\citep{anthropic2024claude35sonnet} and GPT-4o~\citep{achiam2023gpt}; unless stated otherwise, reported numbers use a matched Claude 3.5 Sonnet backbone, so that differences reflect framework design rather than the underlying model.

\paragraph{Knowledge-base and benchmark distributional overlap.} Both the \fa~knowledge base and the FoamBench-Basic cases derive from the OpenFOAM tutorial corpus, so a distributional overlap exists and must be discussed explicitly. Three points bound its effect. First, retrieval returns \emph{structural templates, not solutions}: the agents must still adapt a retrieved tutorial to the requested geometry, boundary conditions, and physical parameters, and the Reviewer must still close the residual gap. The ablation of \Cref{sec:ablation} shows that retrieval alone caps $M_{\mathrm{exec}}$ at $57.3\%$ on FoamBench-Basic. Second, the 16 FoamBench-Advanced cases were hand-crafted by the \fb~authors and are \emph{not} present in our knowledge base, so the Advanced-tier results are out-of-distribution (OOD) evidence. Third, the non-tutorial case studies of \Cref{sec:extmesh} and \Cref{sec:burner} (user-supplied geometries and a multi-physics burner) provide further non-tutorial evidence.

\subsubsection{Extended Baseline Comparison}
\label{sec:extbaseline}
\Cref{tab:main_results} reports $M_{\mathrm{exec}}$ and $M_{\mathrm{NMSE}}$ for \fa, MetaOpenFOAM, and the zero-shot LLM baseline on both tiers under the matched-backbone setting. The zero-shot row is the framework-free lower bound and isolates the agentic-orchestration contribution from raw LLM capability: \fa's $M_{\mathrm{exec}}$ is roughly $14\times$ the zero-shot value on Basic and $37\times$ on Advanced, whereas MetaOpenFOAM is $7$--$9\times$. \fa~leads MetaOpenFOAM by $+0.327$ $M_{\mathrm{exec}}$ and $+0.304$ $M_{\mathrm{NMSE}}$ on Basic, and by $+0.500$ and $+0.281$ on Advanced.

\begin{table}[tbp]
\centering
\caption{Execution success ($M_{\mathrm{exec}}$) and bucketed field-level fidelity ($M_{\mathrm{NMSE}}$) on FoamBench under the matched-backbone setting. Higher is better for both metrics.}
\label{tab:main_results}
\begin{tabular}{llcc}
\toprule
System & Tier & $M_{\mathrm{exec}}$ & $M_{\mathrm{NMSE}}$ \\
\midrule
\fa            & Basic ($n{=}110$)   & \textbf{0.882} & \textbf{0.477} \\
\fa            & Advanced ($n{=}16$) & \textbf{0.625} & \textbf{0.406} \\
MetaOpenFOAM   & Basic ($n{=}110$)   & 0.555 & 0.173 \\
MetaOpenFOAM   & Advanced ($n{=}16$) & 0.125 & 0.125 \\
Zero-shot LLM  & Basic ($n{=}110$)   & 0.064 & 0.050 \\
Zero-shot LLM  & Advanced ($n{=}16$) & 0.017 & 0.009 \\
\bottomrule
\end{tabular}
\end{table}

Because the OpenFOAMGPT~\citep{pandey2025openfoamgpt} source code is not publicly available, it cannot be re-run under the FoamBench protocol, and its published evaluation (six hand-picked tutorials scored by a binary success criterion) is not directly portable to our metrics. To approximate its dependency-handling behavior we ran an \emph{OpenFOAMGPT-shaped proxy}, our pipeline with sequential, dependency-aware file generation replaced by parallel generation (no predecessor context), while retaining the multi-index retrieval and the Reviewer, which scores $0.845$ $M_{\mathrm{exec}}$ / $0.409$ $M_{\mathrm{NMSE}}$ on FoamBench-Basic. Full \fa~leads this proxy by $+0.037$ $M_{\mathrm{exec}}$ and $+0.068$ $M_{\mathrm{NMSE}}$, isolating the contribution of dependency-aware generation. Beyond the experimental baselines above, AutoFEA-like systems~\citep{hou2025autofea} and traditional scripting pipelines lack adaptive retrieval, natural-language-to-domain classification, and iterative repair, and are therefore not directly comparable on the agentic-correctness axis this benchmark measures.

\subsection{Overall Performance Comparison}

\begin{figure*}[tbp]
    \centering\includegraphics[width=1.0\textwidth]{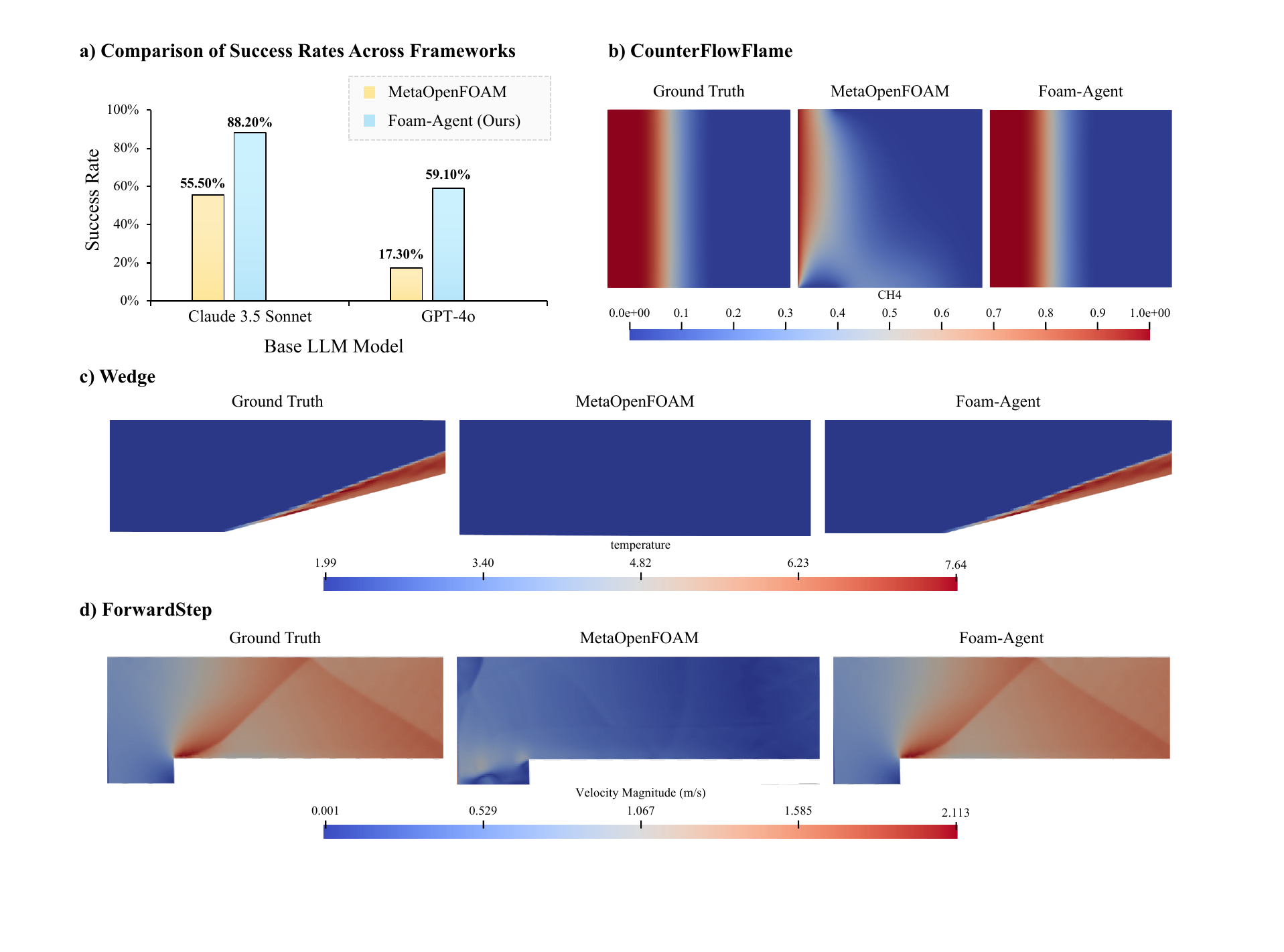} 
    \vspace{-20pt}
    \caption{Performance comparison between MetaOpenFOAM and \fa.
    (a) Execution-success rate ($M_{\mathrm{exec}}$) with Claude 3.5 Sonnet and GPT-4o on the \fb~dataset; the corresponding bucketed field-level fidelity ($M_{\mathrm{NMSE}}$) is reported in \Cref{tab:main_results}.
    (b)--(d) Qualitative comparison of simulation contours against ground truth, complementing the quantitative $M_{\mathrm{NMSE}}$ metric:
    (b) $CH_4$ mass fraction at $t=0.5\,\mathrm{s}$ for the \textit{CounterFlowFlame} case;
    (c) temperature at $t=0.2\,\mathrm{s}$ for the \textit{wedge} case;
    (d) velocity magnitude at $t=4.0\,\mathrm{s}$ for the \textit{forwardStep} case.
    Ground truth is generated by a human expert. \fa~visuals are automatically generated by the agent via Python scripts, which are also used to render the baselines for fair comparison.}
    \label{fig:barplot_with_comparison}
\end{figure*}

\Cref{fig:barplot_with_comparison} and \Cref{tab:main_results} present the comparative performance. \fa~substantially outperforms the baseline across all tested backbones. With Claude 3.5 Sonnet, \fa~reaches an $M_{\mathrm{exec}}$ of $88.2\%$ on FoamBench-Basic, significantly surpassing MetaOpenFOAM's $55.5\%$; the gap is even larger with GPT-4o, where \fa~reaches $59.1\%$ against the baseline's $17.3\%$. Because crash-free execution is not the same as physical correctness, we co-report the bucketed field-level fidelity metric: on FoamBench-Basic \fa~attains $M_{\mathrm{NMSE}}=0.477$, against $0.173$ for MetaOpenFOAM.

\paragraph{Execution success versus solution fidelity.} The headline $88.2\%$ $M_{\mathrm{exec}}$ and the co-reported $M_{\mathrm{NMSE}}=0.477$ are reconciled by the bucket distribution. The 110 Basic-tier cases distribute as $46/13/51$ over the fidelity buckets $\{1,\,0.5,\,0\}$, so the \emph{strict} complete-success rate is $46/110 = 41.8\%$, still well above the upper bound that MetaOpenFOAM's $M_{\mathrm{NMSE}}=0.173$ places on its own complete-success rate. The $51$ bucket-$0$ cases decompose into two distinct failure modes: $13$ are execution failures (NMSE undefined, scored $0$ by convention), while the remaining $38$ \emph{ran to completion but produced a field-level NMSE above $0.3$}. This exec-pass / fidelity-fail population is the operative residual gap; it is analyzed case by case in the failure taxonomy of \Cref{sec:failuretax}, where it is found to be dominated by solver- and parameter-setup errors rather than by file-generation or execution faults.

To complement these aggregate metrics with a qualitative view, we visually compare the outputs against ground-truth solutions for three representative cases from FoamBench: \textit{CounterFlowFlame}, \textit{wedge}, and \textit{forwardStep} (\Cref{fig:barplot_with_comparison}b--d). In the \textit{CounterFlowFlame} case (\Cref{fig:barplot_with_comparison}b), which depicts the CH$_4$ mass fraction at $t=0.5\,\mathrm{s}$, \fa~accurately reproduces the sharp concentration gradients characteristic of flame fronts, whereas the baseline yields a diffuse, inaccurate transition region. Similarly, for the \textit{wedge} case (\Cref{fig:barplot_with_comparison}c), \fa~captures the correct temperature field near the angled wall, while the competing framework fails to reconstruct even the fundamental geometry of the domain. Finally, in the \textit{forwardStep} scenario (\Cref{fig:barplot_with_comparison}d), the velocity-magnitude field generated by \fa~is virtually indistinguishable from the ground truth, whereas the baseline predicts erroneously low velocities throughout the domain. These results underscore the framework's capability to simulate canonical problems with high precision.

\subsection{Ablation Studies}
\label{sec:ablation}
\begin{figure*}[htbp]
    \centering
    \includegraphics[width=1.0\textwidth]{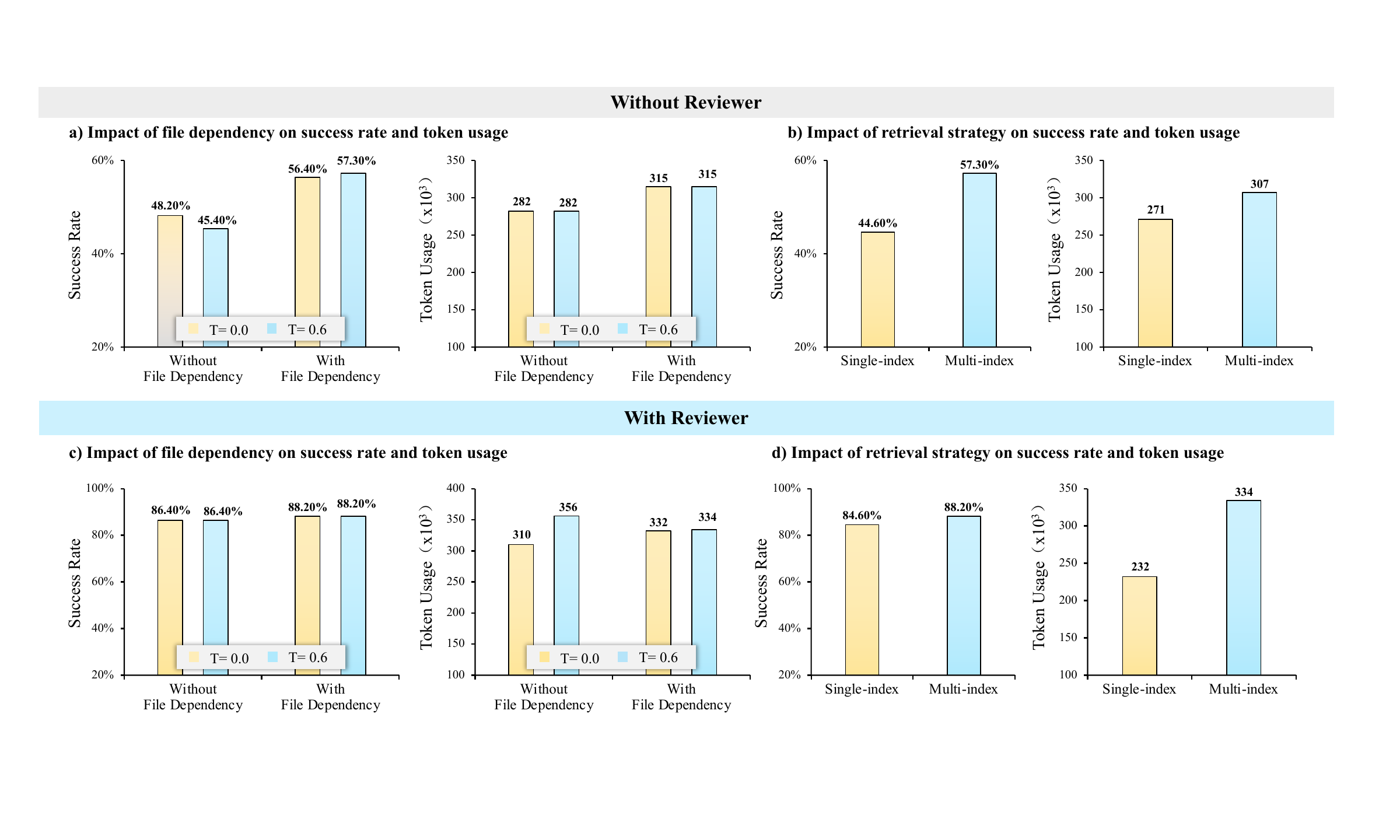}
    \caption{Ablation study of strategies of file generation, retrieval and iterative reviewer used in \fa. We also study the effect of LLM temperature parameters by setting it to a low value of 0 and high value of 0.6.
(a) Impact of file dependency on execution success rate and token usage without the reviewer agent, for LLM temperature values of 0 and 0.6. (b) Impact of retrieval strategy (multi-index versus baseline single-index RAG) on execution success rate and token usage without the reviewer agent, for LLM temperature values of 0 and 0.6. (c) Impact of file dependency on execution success rate, token usage, and average reviewer loops with the reviewer agent, for LLM temperature values of 0 and 0.6. (d) Impact of retrieval strategy (multi-index versus baseline single-index RAG) on execution success rate and token usage with the reviewer agent, for LLM temperature values of 0 and 0.6. Claude 3.5 Sonnet is the prompt model here.}
    \label{fig:ablation_plots}
\end{figure*}

Having shown that \fa~outperforms the baseline, we examine the contribution of each component, focusing on two components: the Reviewer and dependency-aware file generation. Dependency-aware generation orders file creation so that base files precede the dependent files that reference them (\Cref{fig:overview}). All ablation experiments use Claude 3.5 Sonnet~\citep{anthropic2024claude35sonnet}, and \Cref{fig:ablation_plots} summarizes the results. The Reviewer is the single most important factor: it raises the execution-success rate from roughly $50\%$ to over $80\%$ across all configurations. This points to iterative, execution-grounded self-correction~\citep{shinn2023reflexion} as the decisive mechanism. We repeat the experiment at two LLM temperatures ($T$) to confirm that this holds. With the Reviewer disabled, dependency-aware generation gives the largest single improvement in execution success: from $48.2\%$ to $56.4\%$ at $T=0.0$ and from $45.4\%$ to $57.3\%$ at $T=0.6$. With the Reviewer enabled the difference is small, because the loop corrects most initial-generation errors regardless; the main benefit of dependency-aware generation is then faster convergence: it lowers the mean Reviewer-iteration count (from $0.90$ to $0.79$ at $T=0.0$, and from $1.87$ to $0.96$ at $T=0.6$) and thus the API cost and runtime.

The ablation study on \fa's RAG system is summarized in \Cref{fig:ablation_plots}. We first performed an experiment without the reviewer to isolate the effect of multi-index retrieval, which uses a multi-index of case metadata (e.g., case name, domain, category, solver) to retrieve precise context. This multi-index approach outperforms single-index approaches: it reduces noise and improves retrieval precision, narrowing the semantic gap between natural language and technical terminology.

\paragraph{Reviewer $\times$ RAG ablation.} To isolate the structural-retrieval signal from the iterative-repair loop, we ran a $2{\times}2$ ablation on FoamBench-Basic ($n{=}110$) crossing the Reviewer (on/off) with the RAG variant (flat single index versus multi-index); \Cref{tab:ablation2x2} reports the result. Under Reviewer-off, the multi-index lifts $M_{\mathrm{exec}}$ by $+12.8$ percentage points ($44.5\%\to57.3\%$) over the flat baseline, a direct measure of the structural-decomposition contribution, independent of the repair loop. Under Reviewer-on this narrows to $+3.7$ points ($84.5\%\to88.2\%$), consistent with the repair loop absorbing most retrieval-quality variance. Retrieval and iterative repair are thus complementary: retrieval alone caps $M_{\mathrm{exec}}$ at $57.3\%$, and the Reviewer is what carries the system to $88.2\%$.

\begin{table}[tbp]
\centering
\caption{$2{\times}2$ Reviewer $\times$ RAG ablation on FoamBench-Basic ($n{=}110$, LLM temperature $T{=}0.6$).}
\label{tab:ablation2x2}
\begin{tabular}{llc}
\toprule
Reviewer & RAG variant & $M_{\mathrm{exec}}$ \\
\midrule
off & single index & 0.445 \\
off & multi-index  & 0.573 \\
on  & single index & 0.845 \\
on  & multi-index  & \textbf{0.882} \\
\bottomrule
\end{tabular}
\end{table}

\subsubsection{Retrieval Case Studies}
\label{sec:retrievalcases}
Two concrete cases, each run with ten LLM-paraphrased prompts, illustrate the two regimes the ablation aggregates. \emph{Successful retrieval} (\texttt{BenardCells}, a buoyancy-driven natural-convection case): under Reviewer-off the flat single index pulled a structurally adjacent tutorial whose patches were typed as plain \texttt{wall}, so every paraphrase aborted with \texttt{patch type \textquotesingle wall\textquotesingle{} not constraint type}; the multi-index instead surfaced the buoyancy-flagged \texttt{hotRoom} convection tutorial directly, and the Input Writer wrote matching constraint-type patches on the first call ($0/10\to10/10$). Here one structural-retrieval decision replaced an entire Reviewer loop's worth of repair. \emph{Retrieval mismatch recovered by the Reviewer} (\texttt{damBreakWithObstacle}, \texttt{interFoam}): the multi-index retrieved the closest volume-of-fluid tutorial, but its file inventory lacked \texttt{momentumTransport} and \texttt{phaseProperties}, so the first pass tripped a \texttt{cannot find file} error; the Reviewer parsed the missing-dictionary error, mapped it through the dependency graph, and regenerated each missing file, so all ten paraphrases reached the \texttt{End} marker (Reviewer-off $0/10\to$ Reviewer-on $10/10$). Together these cases show retrieval quality and the repair loop as complementary mechanisms.

\subsubsection{Reviewer Convergence and Iteration Count}
\label{sec:convergence}
Because the Reviewer dominates performance, we characterize its dynamics on FoamBench-Basic ($n{=}110$, iteration cap $10$, loop index $0$-indexed, consistent with the reviewer-loop counts plotted in \Cref{fig:ablation_plots}, where loop~$0$ is a case that needs no Reviewer correction). \emph{Iteration count.} The distribution is heavily front-loaded: $74$ cases finish at loop~$0$, $17$ within loops $1$--$2$, $16$ within $3$--$5$, $2$ within $6$--$8$, and $1$ hits the cap; median $0$, mean $0.96$. Thus $67.3\%$ of cases finish at the first loop and the cap-hit rate is only $0.9\%$: in iteration count the Reviewer is a long-tail safety net, even though it dominates wall-clock because each loop brackets a full OpenFOAM run (\Cref{sec:cost}). \emph{Convergence versus oscillation.} Measuring the rewrite-plan size $|\Delta\mathcal{S}_t|$ at each loop and classifying consecutive pairs as oscillation ($|\Delta\mathcal{S}_t|>|\Delta\mathcal{S}_{t-1}|$), stagnation ($|\Delta\mathcal{S}_t|=|\Delta\mathcal{S}_{t-1}|$), or decrease ($|\Delta\mathcal{S}_t|<|\Delta\mathcal{S}_{t-1}|$), we find that, among cases with at least two Reviewer iterations, $47.8\%$ on Basic and $66.7\%$ on Advanced contain at least one oscillation step yet still converge under the cap. The empirical pattern is therefore \emph{closing-by-displacement} rather than monotone error-set decrease: each fix can expose a previously masked downstream error from a later OpenFOAM stage, consistent with the trajectory-conditioned reviewer framing of \Cref{sec:reviewer}.

\subsubsection{Failure Taxonomy}
\label{sec:failuretax}
To locate the residual gap we label every Reviewer-off failure on FoamBench-Basic ($n{=}110$ per RAG arm) with a ten-category symptom taxonomy, recording each category's \emph{origin} (reasoning, tool-chain, or numerical) and its \emph{Reviewer recoverability}, the fraction of those failures recovered by the Reviewer-on rerun of the same case (\Cref{tab:failuretax}). Three readings stand out. First, multi-index retrieval rescues reasoning errors upstream of the Reviewer: the no-Reviewer failure pool shifts from $29$ reasoning / $24$ tool-chain / $8$ numerical under the single index to $10/30/7$ under the multi-index, and category~3 (boundary-condition name mismatch) collapses from $10$ cases to $0$. Second, the Reviewer is highly effective on dictionary-grade errors: it recovers $100\%$ of syntax and boundary-condition errors and $95.5\%$ of missing-file errors, the largest single class. Third, the hardest residual class is solver/model mismatch (category~5, dominated by \texttt{mpirun} option errors), where recovery is only $0$--$12.5\%$ because this runner-script-level evidence is not diagnosable from the solver log seen by the extractor $\Phi$ (\Cref{sec:runner}). Aggregate Reviewer recovery on initial failures is $80.9\%$ under the multi-index; the remaining $\approx19\%$ defines the current diagnostic frontier. This taxonomy is also the lens for the $38$ exec-pass / fidelity-fail cases identified in the overall comparison: they pass $M_{\mathrm{exec}}$ but, dominated by categories~5 and~6, fall short on field-level fidelity.

\begin{table*}[tbp]
\centering
\caption{Ten-category taxonomy of Reviewer-off failures on FoamBench-Basic ($n{=}110$ per RAG arm). ``cnt'' is the number of Reviewer-off failures in that category under the single-index / multi-index arm; ``rec\%'' is the fraction of those failures recovered by the Reviewer-on rerun. ``---'' marks an empty category.}
\label{tab:failuretax}
\setlength{\tabcolsep}{4pt}
\begin{tabular}{cllcccc}
\toprule
\# & Category & Origin & cnt (single) & cnt (multi) & rec\% (single) & rec\% (multi) \\
\midrule
1  & Syntax error                                   & reasoning  &  7 &  7 & 100.0 & 100.0 \\
2  & Missing file                                   & tool-chain & 16 & 22 &  93.8 &  95.5 \\
3  & Dependency inconsistency (BC name)             & reasoning  & 10 &  0 & 100.0 & ---   \\
4  & BC error (wrong type for regime)               & reasoning  & 11 &  3 & 100.0 & 100.0 \\
5  & Solver/model mismatch (incl.\ \texttt{mpirun}) & tool-chain &  8 &  8 &   0.0 &  12.5 \\
6  & Convergence failure (FPE / thermo bounds)      & numerical  &  8 &  7 &  50.0 &  85.7 \\
7  & Mesh quality (\texttt{checkMesh})              & tool-chain &  0 &  0 & ---   & ---   \\
8  & Geometry/meshing mismatch                      & reasoning  &  1 &  0 &   0.0 & ---   \\
9  & Unsupported workflow (LES / overset)           & tool-chain &  0 &  0 & ---   & ---   \\
10 & Unrecoverable reasoning (catch-all)            & reasoning  &  0 &  0 & ---   & ---   \\
\midrule
\multicolumn{3}{l}{Total} & 61 & 47 & 77.0 & 80.9 \\
\bottomrule
\end{tabular}
\end{table*}

\begin{figure*}[htbp]
    \centering
    \includegraphics[width=1.0\textwidth]{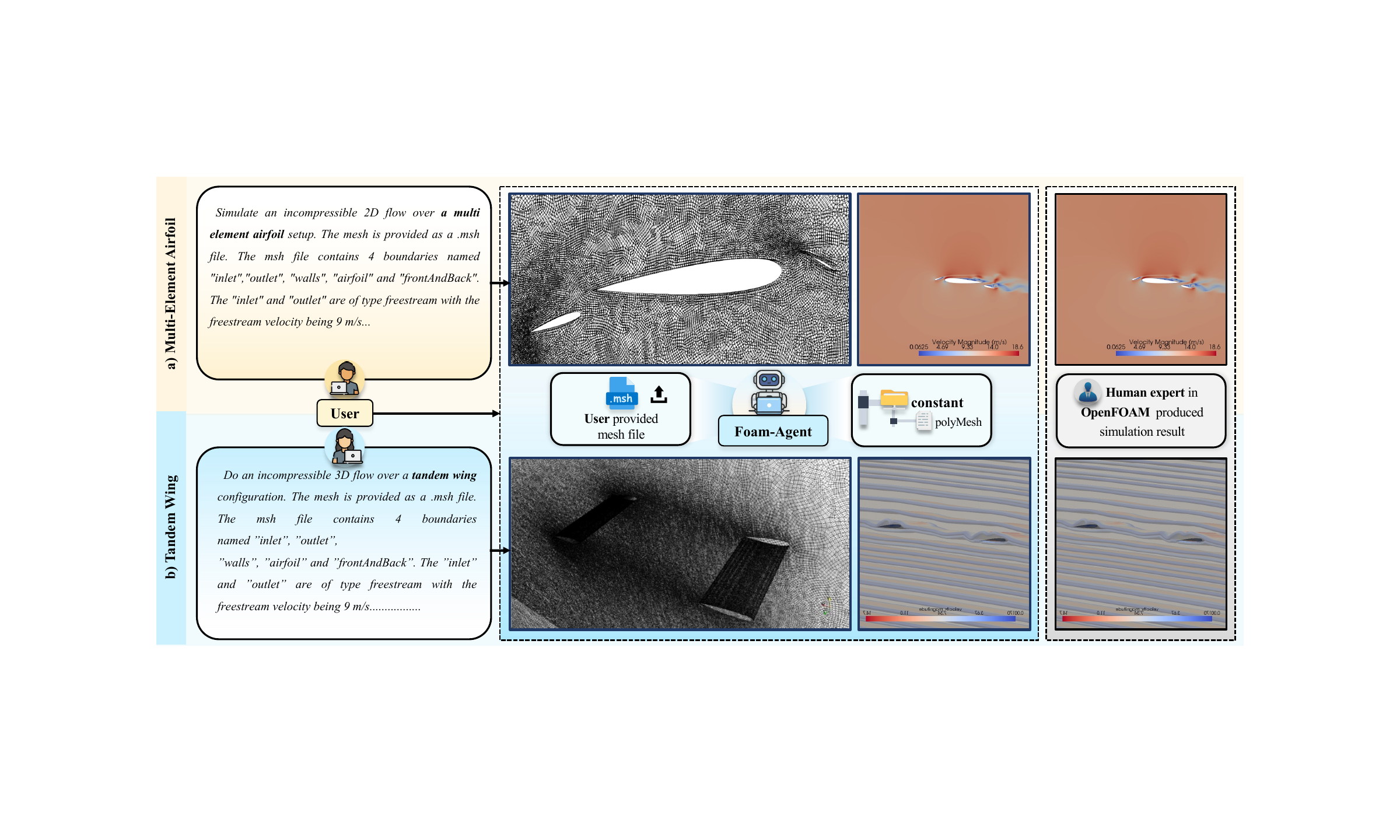}
    \caption{External mesh cases. \fa converts user-supplied \texttt{.msh} files to OpenFOAM format, generates the remaining case files, and produces final velocity-magnitude visualizations for (a) a multi-element airfoil and (b) a tandem wing. The same visualization script is used for the agent and expert solutions to ensure a consistent qualitative comparison.}
    \label{fig:mesh_upload}
\end{figure*}

\begin{figure*}[!htbp]
    \centering
    \includegraphics[width=1.0\textwidth]{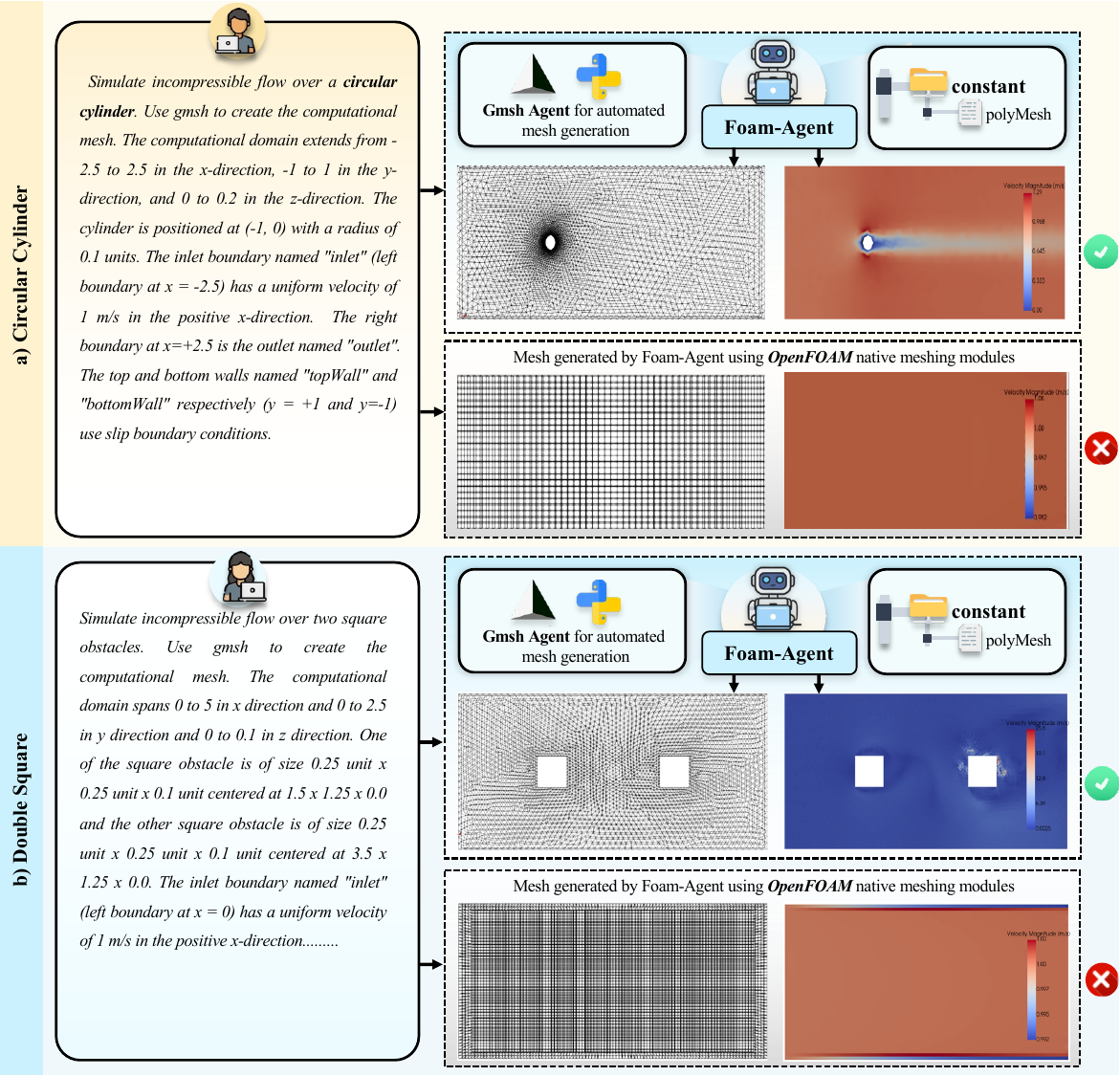}
    \caption{Tool-based mesh generation with Gmsh. For (a) flow over a cylinder and (b) flow over two square obstacles, \fa generates a Gmsh Python script, converts the resulting \texttt{.msh} file to OpenFOAM format, runs the case, and visualizes the final velocity magnitude. OpenFOAM native meshing does not represent the obstacles correctly in these examples.}
    \label{fig:gmsh}
\end{figure*}

\subsection{External Mesh}
\label{sec:extmesh}
Processing externally developed mesh files is a \emph{supported capability} of \fa: it is a deliberate human--AI design boundary that lets a user supply an expert mesh when the in-the-loop meshing tools fall short for a given geometry. The dominant added bottleneck of this compute form is mesh conversion and boundary-patch consistency; \fa~handles it by converting the supplied mesh to the OpenFOAM \texttt{polyMesh} format, verifying the patch inventory, and escalating to patch renaming or plan revision when the patch names do not match the prompt. Meshes are provided as \texttt{.msh} files, with the boundary names and flow scenario described in the prompt. We demonstrate this capability on two cases:
1) Multi-element airfoil~\citep{tandem_wing_repo}: This case describes the flow around multiple airfoils within the domain, with the flow of one affecting the flow of another. This 2D simulation uses an inlet velocity of 9 m/s and a fluid kinematic viscosity of $1.5\times10^{-5}$ $\mathrm{m^2/s}$. The simulation is set to use the \texttt{simpleFoam} solver and Spalart-Allmaras turbulence model~\citep{spalart1992one}, with a timestep of 1.0 s and a final time of 1000 s.
2) Tandem wing configuration~\citep{tandem_wing_repo}: This case describes the flow around a tandem configuration, with one wing located in the wake of the other. This 3D simulation uses an inlet velocity of 9 m/s and a fluid kinematic viscosity of $1.5\times10^{-5}$ $\mathrm{m^2/s}$. The simulation is set to use the \texttt{simpleFoam} solver and the Spalart-Allmaras turbulence model, with a timestep of 1.0 s and a final time of 1000 s. The prompts along with the path to the mesh are provided to \fa~for simulating the flow field; the complete user prompts for this and the other case studies are reproduced in \Cref{app:b}. The corresponding mesh and the contour of the velocity magnitude at the final timestep for these cases is shown in~\Cref{fig:mesh_upload}. Further, we also compare \fa~results with human expert-generated simulation results. The simulation results from \fa~closely match with the result from a human OpenFOAM expert as shown in the velocity magnitude contour plots in~\Cref{fig:mesh_upload}. Inspecting the case files generated by the agent against those prepared by humans, we found them to closely match in key aspects such as physical parameters, turbulence model selection, boundary condition assignment, and solver choice. Minor differences were observed in the numerical schemes, but these did not significantly affect the outcomes, as confirmed by the velocity contours.

\subsection{Tool-Based Mesh Generation}
\label{sec:gmsh}
We demonstrate the mesh-generation capabilities of \fa~using the Gmsh~\citep{geuzaine2009gmsh} Python library on two cases: 1) Flow over a cylinder. 2) Flow over two square obstacles. The simulation is run for 10~s using the \texttt{pisoFoam}~\citep{jasak2007openfoam} solver. The mesh generated by the agent and the contour of velocity magnitude at the final timestep for the two cases is shown in \Cref{fig:gmsh}. To underscore the necessity of a specialized meshing agent that leverages external tools such as Gmsh, we compare meshes generated with OpenFOAM's native meshing utilities against those produced via the Gmsh Python API. The native meshing modules were unable to capture the intended geometry of the flow scenario, whereas the Gmsh-based approach successfully generated the overall domain and accurately represented the obstacles within it. The meshing agent creates the mesh in the form of \texttt{.msh} file, which is then converted to OpenFOAM compatible format and then further used for simulation.

The dominant added bottleneck of the Gmsh pathway is geometry-to-\texttt{.geo} generation together with Gmsh syntax or geometry repair and the \texttt{.msh}$\to$\texttt{polyMesh} conversion; \fa~handles it by having the Meshing Agent generate and repair the \texttt{.geo} script, with Reviewer iterations re-entering the meshing loop whenever a conversion or patch check fails. A per-mesh-type reliability breakdown on the FoamBench-Advanced tier is reported separately in \Cref{app:suppl}.

\subsection{HPC Execution}
\label{sec:hpc}
\begin{figure*}[tbp]
    \centering
    \includegraphics[width=1.0\textwidth]{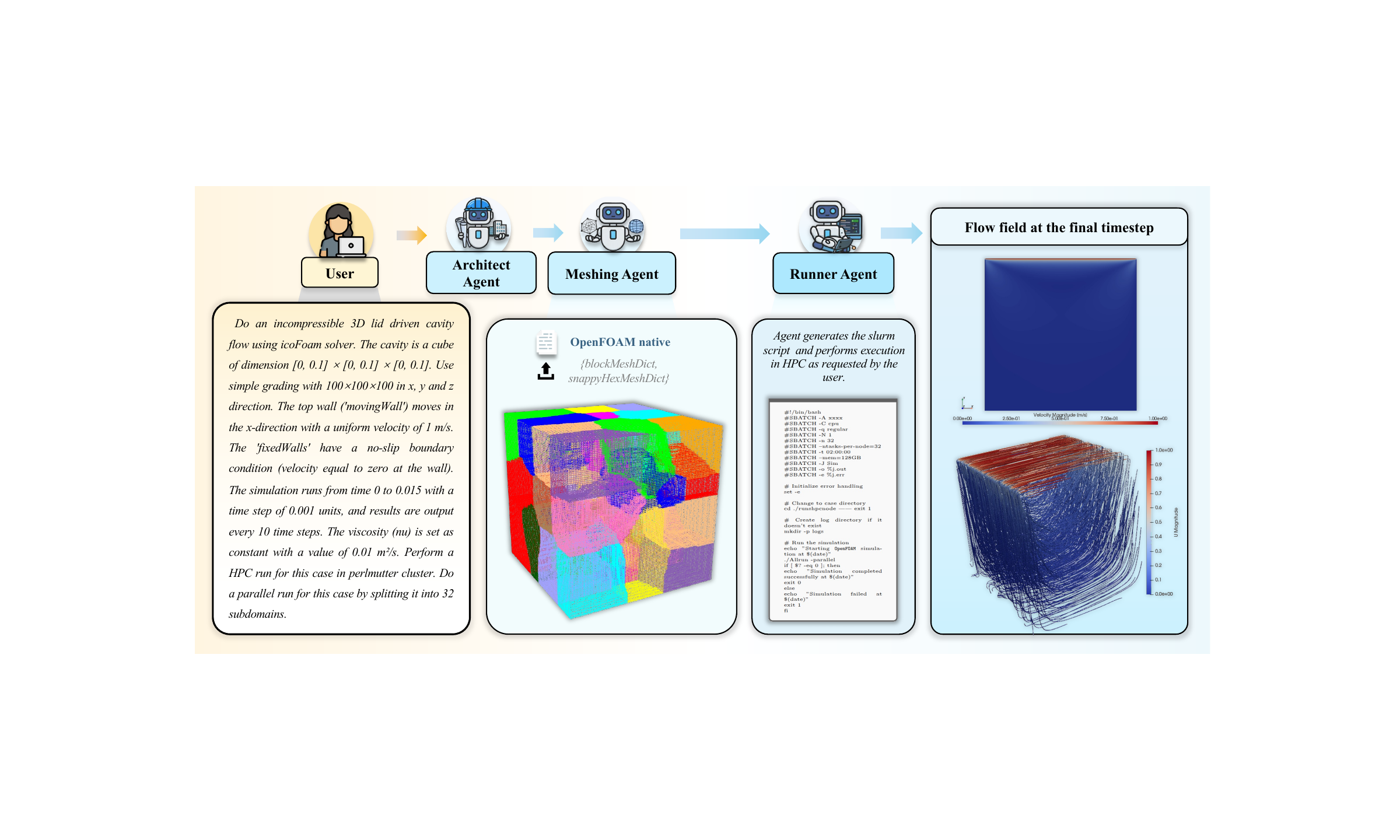}
    \caption{HPC execution of a 3D lid-driven cavity case. \fa generates the OpenFOAM setup, domain decomposition, and Slurm script for Perlmutter, runs the case on 32 subdomains, and produces the final mid-\textit{z} velocity contour. The 3D streamlines are included only for qualitative illustration.}
    \label{fig:cavity_hpc}
\end{figure*}

We demonstrate the capabilities of the HPC Runner Agent by instructing the framework to perform a 3D lid-driven cavity flow with one million cells, following the setup used in~\citep{openfoam_hpc_tc2025}. The agent generates the necessary OpenFOAM case files along with a Slurm submission script for the HPC platform \textit{Perlmutter}~\citep{nersc_perlmutter}. The full Slurm script produced by the agent is reproduced in \Cref{app:c}.
To ensure strict adherence to platform-specific scheduling policies, which often vary significantly across clusters, we adopt a documentation-guided approach.
Instead of relying solely on the LLM's parametric knowledge, relevant documentation regarding partition limits, module naming conventions, and header syntax is injected into the agent's context (via RAG or direct prompting). 
This context-aware generation reduces the risk of hallucinated directives and improves the reliability of the generated executable scripts.
The dominant added bottleneck of HPC execution is not introduced by \fa~but inherited from cluster computing at large: scheduler and queue latency, job stage-in and stage-out time, and allocation-specific configuration; \fa~confines the agent's responsibility to generating a valid Slurm script, while the user-supplied partition, account, node-count, and wall-time fields define the human--AI boundary.
The resulting highly refined mesh, the velocity contours at the final timestep (mid-$z$ slice), and corresponding streamlines are presented in \Cref{fig:cavity_hpc}.

\subsection{Computational Cost}
\label{sec:cost}
We characterize the computational cost of \fa~along four dimensions: end-to-end runtime against a human-expert baseline, per-agent wall-clock attribution, token and dollar cost, and scalability with problem size.

\paragraph{Runtime versus manual setup.} Three standard cases were independently set up by a senior OpenFOAM user (about five years' experience) under a controlled protocol: start from the closest OpenFOAM~v10 tutorial, consult only the OpenFOAM user guide (no LLM tools), and meet the same complete-success criterion as \fa. The human time covers code writing, review, and iterative editing but excludes the solver wall-clock incurred between iterations; the \fa~time is the measured end-to-end wall-clock and \emph{includes} solver execution between Reviewer iterations, so the speed-up of \Cref{tab:runtime} is conservative on the \fa~side. \fa~is $2.1$--$14.5\times$ faster than the human expert across the three cases.

\begin{table}[tbp]
\centering
\caption{End-to-end runtime of \fa~versus a senior OpenFOAM user on three standard cases.}
\label{tab:runtime}
\begin{tabular}{lccc}
\toprule
Case & Human (min) & \fa~(min) & Speed-up \\
\midrule
Lid-driven cavity (\texttt{icoFoam})           & 16 & 1.1 & $14.5\times$ \\
Cylinder flow (incompressible)                 & 34 & 3.9 & $8.7\times$ \\
\texttt{forwardStep} (\texttt{rhoCentralFoam}) & 15 & 7.3 & $2.1\times$ \\
\bottomrule
\end{tabular}
\end{table}

\paragraph{Per-agent wall-clock.} \Cref{tab:peragent} attributes end-to-end wall-clock to each agent, computed from LLM-call timestamps. The Reviewer step dominates ($74.7\%$ of attributed time on FoamBench-Basic and $83.7\%$ on Advanced) because each Reviewer pass brackets a full OpenFOAM solver run between writing the patched files and the next error-analysis call; the Architect, Input Writer, and Runner together account for the remainder. Per-case end-to-end wall-clock is $370$\,s mean / $148$\,s median on Basic and $714$\,s mean / $302$\,s median on Advanced. The Mesh and Visualization agents do not appear in this trace: meshing shells out to \texttt{blockMesh}, \texttt{snappyHexMesh}, or Gmsh without an LLM call, and visualization runs locally.

\begin{table}[tbp]
\centering
\caption{Per-agent wall-clock attribution on FoamBench, from LLM-call timestamps. The Reviewer share includes the OpenFOAM solver execution it brackets.}
\label{tab:peragent}
\begin{tabular}{lcccc}
\toprule
& \multicolumn{2}{c}{Basic} & \multicolumn{2}{c}{Advanced} \\
\cmidrule(lr){2-3}\cmidrule(lr){4-5}
Agent & \% & mean (s) & \% & mean (s) \\
\midrule
Reviewer     & 74.7 & 276.4 & 83.7 & 597.7 \\
Input Writer & 11.9 &  44.1 &  7.4 &  53.1 \\
Architect    &  7.1 &  26.2 &  5.7 &  40.8 \\
Runner       &  6.3 &  23.3 &  3.1 &  22.3 \\
\midrule
Total        & ---  & 370.0 & ---  & 713.9 \\
\bottomrule
\end{tabular}
\end{table}

\paragraph{Token and dollar cost.} \Cref{tab:cost} reports token usage and dollar cost at a rate of \$3 per million input tokens and \$15 per million output tokens. The Basic-tier median cost (\$0.29 per case) sits well below the mean (\$1.07), a gap driven by a long tail of multi-iteration cases (maximum \$8.97 per case), consistent with the long-tailed Reviewer-iteration distribution of \Cref{sec:convergence}.

\begin{table}[tbp]
\centering
\caption{Token usage and dollar cost on FoamBench, at \$3/M input and \$15/M output tokens.}
\label{tab:cost}
\begin{tabular}{lcccccc}
\toprule
Tier & Cases & Calls & Input & Output & Mean \$ & Median \$ \\
\midrule
Basic    & 110 & 1\,928 & 36.13\,M & 0.65\,M & 1.07 & 0.29 \\
Advanced &  16 &   557  &  5.20\,M & 0.40\,M & 1.35 & 1.27 \\
\bottomrule
\end{tabular}
\end{table}

\paragraph{Scalability with problem size.} Because the six agents operate on configuration files (\texttt{system/}, \texttt{constant/}, \texttt{0/}, \texttt{Allrun}) rather than on mesh data, their LLM-call time is independent of mesh resolution by construction. Solver wall-clock, in contrast, follows the standard CFD cost model (linear in cell count for explicit schemes, super-linear for implicit). Agent overhead is therefore a \emph{shrinking} fraction of end-to-end cost as the problem grows. In this regime each saved Reviewer iteration avoids a full expensive run, so dependency-aware generation (which empirically lowers the iteration count) becomes disproportionately valuable on large cases. A full stratified HPC scaling grid is left as a dedicated benchmarking study.

\subsection{Multi-Physics Case Study: A Multi-Region Burner}
\label{sec:burner}
The FoamBench-Advanced tier already exercises advanced turbulence (LES, Spalart--Allmaras, $k$--$\varepsilon$, $k$--$\omega$ SST), complex geometries, and partially multi-physics cases. To complement that breadth with an in-depth multi-physics walkthrough, we present a \emph{multi-region burner} case that \fa~configured and ran end-to-end from a natural-language description, with no manual intervention to the generated case files.

\paragraph{Case description.} The case is an axisymmetric, two-region conjugate-heat-transfer combustion chamber solved with \texttt{chtMultiRegionFoam} (OpenFOAM~10). A gas region and a surrounding solid wall region exchange heat across their shared interface, representative of the thermal loading of industrial furnace walls. Methane enters through a central fuel inlet and air through an annular co-flow inlet, and combustion products exit downstream. Chemical kinetics use the single-step Westbrook--Dryer mechanism integrated with an Euler-implicit stiff ODE solver, and turbulence--chemistry interaction is handled by the Partially Stirred Reactor (PaSR) model. The turbulence closure is $k$--$\omega$ SST, thermodynamic properties are computed from JANAF polynomials with a compressible perfect-gas equation of state, and thermal radiation is modeled with the P1 approximation. The case therefore couples reactive multi-species flow, turbulence--chemistry interaction, conjugate heat transfer, thermal radiation, and compressible thermodynamics, a combination that demands expert-level configuration in a standard OpenFOAM workflow.

\begin{figure*}[tbp]
    \centering
    \begin{minipage}[b]{0.48\linewidth}
        \centering
        \includegraphics[width=\linewidth]{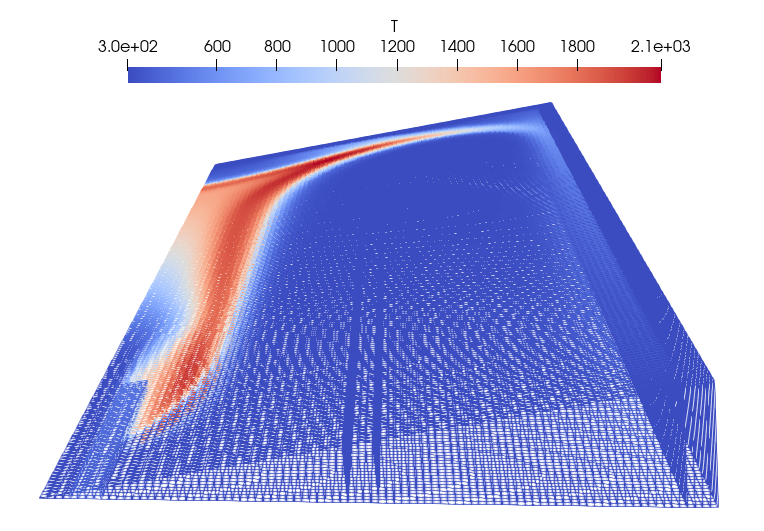}\\
        \small (a) Expert reference
    \end{minipage}
    \hfill
    \begin{minipage}[b]{0.48\linewidth}
        \centering
        \includegraphics[width=\linewidth]{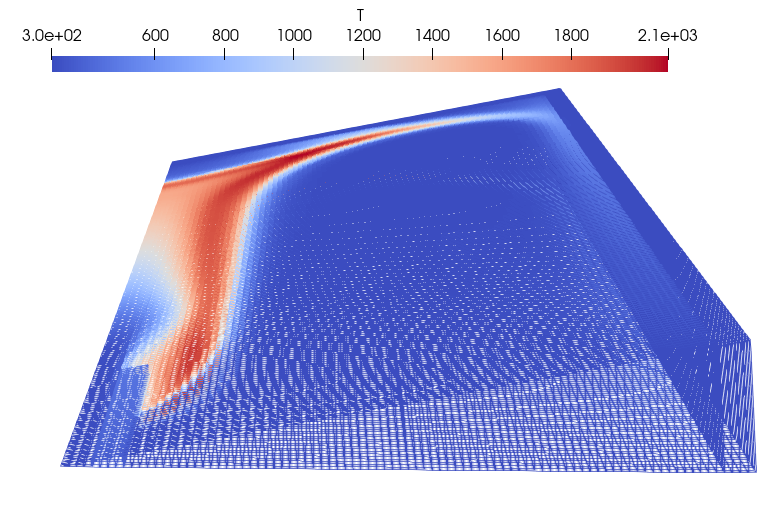}\\
        \small (b) \fa
    \end{minipage}
    \caption{Steady-state temperature field $T$ [K] for the multi-region burner case (\texttt{chtMultiRegionFoam}; $k$--$\omega$ SST $+$ PaSR $+$ Westbrook--Dryer kinetics $+$ conjugate heat transfer $+$ P1 radiation). (a)~Expert reference configured by a senior OpenFOAM user; (b)~\fa~configured autonomously from a natural-language description. Both resolve a $\approx2100$\,K flame zone anchored at the fuel inlet; the field-level NMSE between (a) and (b) at the final time step is $0.0398$ ($M_{\mathrm{NMSE}}{=}1$).}
    \label{fig:reverse_burner}
\end{figure*}

\paragraph{Result.} \Cref{fig:reverse_burner} compares the steady-state temperature field of an expert reference simulation against the case generated autonomously by \fa. Both resolve the core flame structure: a high-temperature reaction zone ($T\approx2100$\,K) anchored near the fuel inlet that expands radially into the chamber, with the surrounding gas near the inlet temperature ($T\approx300$\,K). The field-level NMSE between the two solutions at the final time step is $0.0398$, placing the case in the top fidelity bucket ($M_{\mathrm{NMSE}}{=}1$) and confirming the qualitative agreement quantitatively. \fa~thus navigated the full multi-physics setup (reactive flow, conjugate heat transfer, radiation, and compressible thermodynamics) without expert oversight.

\subsection{Interoperability demonstration}
\begin{figure*}[htbp]
    \centering
    \includegraphics[width=1.0\textwidth]{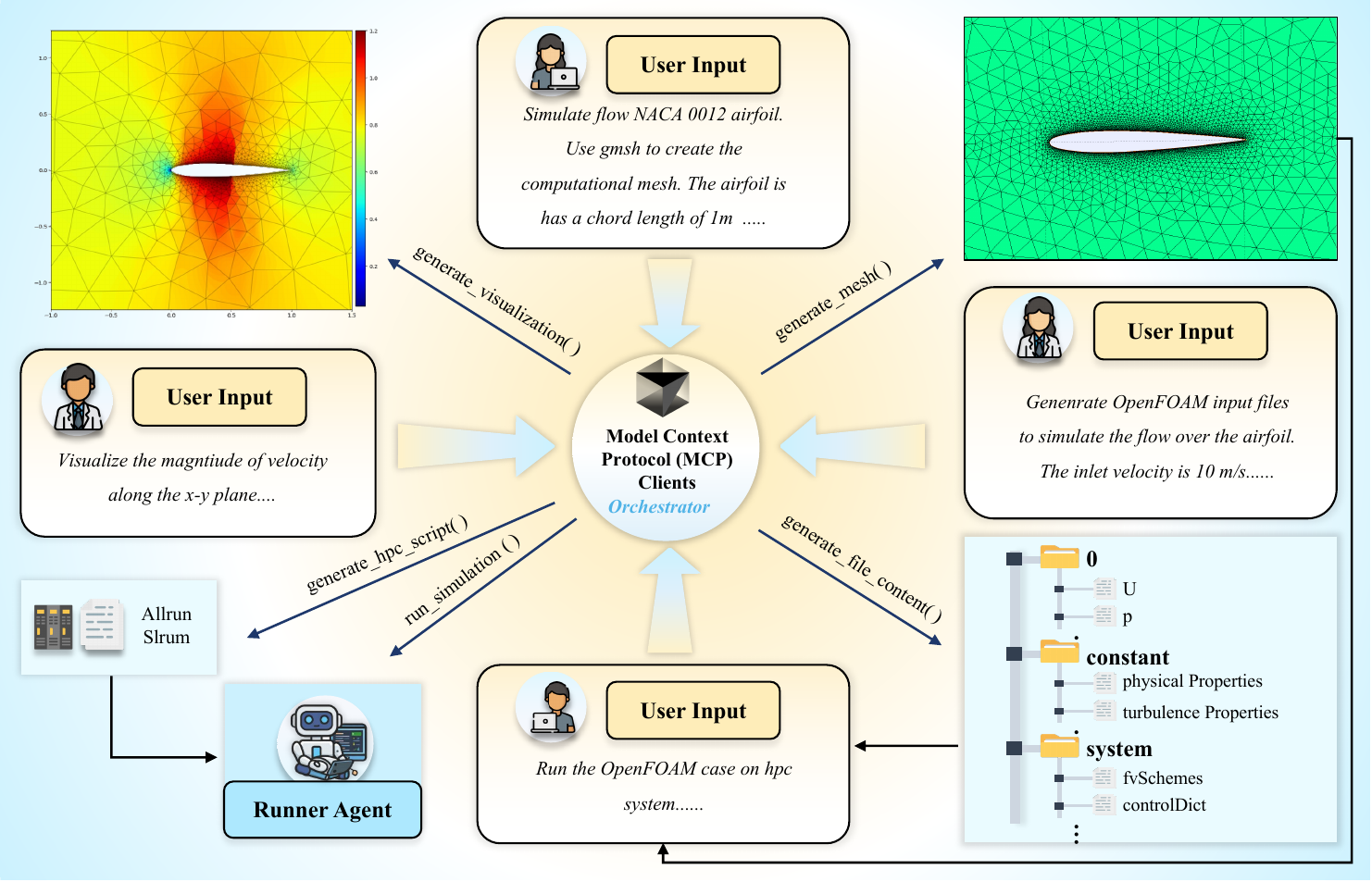}
    \caption[MCP-orchestrated NACA0012 workflow]{MCP-orchestrated end-to-end simulation of NACA0012 flow. The orchestrator invokes \texttt{generate\_mesh()}, \texttt{generate\_file\_content()}, \texttt{generate\_hpc\_script()}, \texttt{run\_simulation()}, and \texttt{generate\_visualization()} to complete the workflow from natural-language request to final result.}
    \label{fig:mcp_example}
\end{figure*}

To show that \fa~can be driven as a component rather than only as a standalone pipeline, we expose its agent functions over MCP and let an external orchestrator sequence them. 
The MCP server design is shown in \Cref{fig:mcp} (\Cref{app:d}). 
In this study, we utilize the cursor environment as the orchestrator to perform a complete end-to-end simulation of the flow over a NACA~0012 airfoil (\Cref{fig:mcp_example}). 
Specifically, the orchestrator first generates the mesh using Gmsh by invoking~\texttt{generate\_mesh()}. Next, it creates the input files via~\texttt{generate\_file\_content()} (the Input Write agent). 
Once the files are ready, it targets an HPC node by calling~\texttt{generate\_hpc\_script()} to produce~\texttt{Allrun} and Slurm scripts, and then submits the job with~\texttt{run\_simulation()}. 
Finally, it renders the velocity magnitude by invoking~\texttt{generate\_visualization()}, which triggers the Visualization node.

\section{Conclusion}
\label{sec:conclusion}

This work introduced \fa as a domain-structured harness for automating OpenFOAM-based CFD workflows from natural language. The central goal is not merely to generate case files, but to make the latent structure of CFD case construction explicit enough that it can be retrieved, executed, inspected, and repaired. This design is not tied to a particular retrieval backend, language model, or orchestration protocol. Rather, it rests on three methodological components: multi-index retrieval over solver, physics, boundary-condition, and file-level knowledge; dependency-aware file generation through a topological traversal of the OpenFOAM case dependency graph; and a trajectory-conditioned reviewer loop that repairs failed runs through execution-grounded self-correction. Around these components, \fa organizes specialist agents for mesh generation, case setup, execution, debugging, and post-processing, and exposes the resulting workflow through a Model Context Protocol deployment surface for use by external orchestrators.
The results on \fb show that this structure significantly improves automated CFD case construction. \fa achieves execution-success rates of $88.2\%$ on the 110 Basic-tier cases and $62.5\%$ on the out-of-distribution Advanced tier, $1.6\times$ and $5\times$ those of the prior MetaOpenFOAM framework, both with Claude 3.5 Sonnet. At the same time, we also report bucketed field-level fidelity, because crash-free execution is only a necessary condition for useful CFD automation. This separation is important for evaluating scientific agents: execution success measures whether the workflow can produce a runnable simulation, while field-level fidelity tests whether the resulting flow is physically and numerically credible for the intended problem.

The main lesson from these experiments is that many failures in LLM-driven CFD automation are structural rather than purely linguistic. Incorrect simulations often arise from inconsistent dependencies across case files, incompatible solver and boundary-condition choices, insufficiently constrained mesh generation, or local repair actions that fix one error while damaging the global case structure. \fa addresses these issues by combining specialist agents with an explicitly serial dependency structure aligned with OpenFOAM practice. This design provides modularity without allowing unconstrained inter-agent negotiation to dominate a workflow that is naturally ordered by physical and numerical dependencies.
The present system remains limited in several respects. It is most reliable for simple-to-moderate geometries generated through \texttt{blockMesh}, \texttt{snappyHexMesh}, or Gmsh, although it can ingest external expert meshes when needed. Its repair capability is centered on mesh regeneration, mesh-parameter adjustment, execution diagnosis, and case-level correction, rather than closed-loop modification of CAD or the pre-meshing geometry definitions. Finally, the inter-file dependency structure is currently implemented as a fixed topological order derived from the OpenFOAM directory convention, whereas future versions could infer case-specific dependencies and compile them into reusable workflows, directly addressing known multi-file consistency failures in language-model agents for scientific computing~\citep{duston2025ainsteinbench}.

Future efforts should target expanding both the physical fidelity of the computational harness and the cognitive autonomy of the multi-agent architecture.
On the CFD side, code-level modification can enable LLM-driven development of low-level components of CFD solvers. For example, turbulence models for LES~\citep{yang2025llmturbulence}. 
On the application side, real-world OpenFOAM workflow that  requires multi-physics and advanced geometric understanding, such as fluid--structure interaction, overset and sliding meshes, solve-time adaptive mesh refinement, is not fully represented in the current harness. Broader validation on non-tutorial benchmarks, especially industrial geometries and case sets stratified by mesh-generation pathway, is needed to test whether the same design principles still remain effective beyond tutorial-derived examples. It would be interesting to see the role of RAG and behavior of LLM agents when confronting completely unseen industrial grade problems. 
On the LLM side, while user prompt is sufficient to determine the simulation in this work, it should be clarified in practice, as human can be naturally ambiguous about their intent when interacting with LLM~\citep{somasekharan2026sciconvbench}. 
On the agentic side, decentralized organizations may be complementary for less structured open-ended discovery settings~\citep{yue2025autonomous}, while a vision-language model could inspect generated visualizations, compare them with expected physical patterns, and provide result-level feedback~\citep{somasekharan2026ai}. Furthermore, human-agent collaborative paradigms~\citep{shao2026sciscigpt} could enable the system to gracefully solicit human intervention when encountering unrecoverable states, thereby ensuring reliability while preserving a substantial speedup over purely manual baselines. Taken together, these directions position \fa as domain-level CFD harness engineering: a composable and empirically testable workflow harness that distills expert OpenFOAM practice into a reusable component for automated numerical experiments of fluid flows.

\printcredits

\section*{Declaration of competing interests}
The authors declare that they have no known competing financial interests or personal relationships that could have appeared to influence the work reported in this paper.

\section*{Declaration of generative AI and AI-assisted technologies in the manuscript preparation process}
During the preparation of this work the authors used ChatGPT (OpenAI) to assist with LaTeX formatting and language editing. 
After using this tool, the authors reviewed and edited the content as needed and take full responsibility for the content of the published article.

\section*{Data statement}
The datasets generated and analyzed during the current study are available at \url{https://github.com/csml-rpi/Foam-Agent/tree/main/database}. 
The evaluation datasets and associated scripts are based on the \fb framework, which is publicly available at \url{https://github.com/NLR-Theseus/cfdllmbench}. Source data are provided with this paper.

\section*{Code availability}
The complete \fa~code base, together with all system and user prompts, run configurations, evaluation scripts, and execution logs, is publicly available at \url{https://github.com/csml-rpi/Foam-Agent} under the MIT license. This supports full reproduction of the analyses and results reported in this study.

\section*{Acknowledgements}
This work received funding support from the U.S. Department of Energy with ID DE-SC0025425. Shaowu Pan is supported by the Google Research Scholar Program. 
Computing resources are supported by the Lambda research grant program, NSF-ACCESS-PHY240112 and by the National Energy Research Scientific Computing Center under award NERSC DDR-ERCAP0030714.

\bibliographystyle{cas-model2-names}
\bibliography{main}

\appendix

\section{System and User Prompts}
\label{app:a}

This appendix presents all system and user prompts used in the \fa framework for various components. These prompts are organized by agent role and function within the multi-agent architecture.

\subsection{Architect Agent Prompts}

The Architect Agent interprets user requirements into a structured simulation plan and breaks down complex tasks into manageable subtasks.

\subsubsection{Case Description Prompts}

\begin{tcolorbox}[colback=blue!5!white,colframe=blue!75!black,title=Case Description System Prompt]
Please transform the following user requirement into a standard case description using a structured format.
The key elements should include case name, case domain, case category, and case solver.
\end{tcolorbox}

\begin{tcolorbox}[colback=blue!5!white,colframe=blue!75!black,title=Case Description User Prompt]
User requirement: \{user\_requirement\}.
\end{tcolorbox}
user\_requirement here refers to the input simulation request from the user. Example, see~\Cref{fig:overview} in the main manuscript.
\subsubsection{Task Decomposition Prompts}

\begin{tcolorbox}[colback=blue!5!white,colframe=blue!75!black,title=Task Decomposition System Prompt]
You are an experienced Planner specializing in OpenFOAM projects. 
Your task is to break down the following user requirement into a series of smaller, manageable subtasks. 
For each subtask, identify the file name of the OpenFOAM input file (example: U, system etc.) and the corresponding folder name where it should be stored. 
Your final output must strictly follow the JSON schema below and include no additional keys or information:

\{ \\
``subtasks'': [ \\
    \{ \\ 
      ``file\_name'': ``$<$string$>$'', \\ 
      ``folder\_name'': ``$<$string$>$'' \\ 
    \} \\ 
    // ... more subtasks \\
  ] \\
\}

Make sure that your output is valid JSON and strictly adheres to the provided schema.
Make sure you generate all the necessary files for the user's requirements.
\end{tcolorbox}

\begin{tcolorbox}[colback=blue!5!white,colframe=blue!75!black,title=Task Decomposition User Prompt]
User Requirement: \{user\_requirement\}

Reference Directory Structure (similar case): \{dir\_structure\}

\{dir\_counts\_str\}

Make sure you generate all the necessary files for the user requirements.
Please generate the output as structured JSON.
\end{tcolorbox}
dir\_structure refers to the folders and files that need to be created for simulating the requested scenario. For example, \texttt{0/U}, \texttt{constant/momentumTransport}, \texttt{system/controlDict} etc. dir\_counts\_str refers to the number of directories to be created (\texttt{0}, \texttt{constant}, \texttt{system} etc.).

\subsection{Input Writer Agent Prompts}

The Input Writer Agent generates OpenFOAM configuration files and ensures consistency across interdependent files based on the user requirements and the simulation plan obtained from the architect agent. The architect agent provides the information of the files to be written and the folder it has to be saved into to the input writer agent, which writes the content of these files. Below, the instructions for the Input Writer Agent are outlined for generating configuration files, terminal commands, and scripts.

\subsubsection{File Generation Prompts}
\begin{tcolorbox}[colback=blue!5!white,colframe=blue!75!black,title=File Generation System Prompt]
You are an expert in OpenFOAM simulation and numerical modeling.
Your task is to generate a complete and functional file named: $<$file\_name$>$\{file\_name\}$<$/file\_name$>$ within the $<$folder\_name$>$\{folder\_name\}$<$/folder\_name$>$ directory. 
Ensure all required values are present and match with the files content already generated.
Before finalizing the output, ensure: 
\begin{itemize}
\item All necessary fields exist (e.g., if `nu' is defined in `constant/transportProperties', it must be used correctly in `0/U'). 
\item Cross-check field names between different files to avoid mismatches. 
\item Ensure units and dimensions are correct for all physical variables. 
\item Ensure case solver settings are consistent with the user requirements. 
\end{itemize}
Available solvers are: \{state.case\_stats[`case\_solver']\}. Provide only the code -- no explanations, comments, or additional text.
\end{tcolorbox}

\begin{tcolorbox}[colback=blue!5!white,colframe=blue!75!black,title=File Generation User Prompt]
User requirement: \{state.user\_requirement\}. 
Refer to the following similar case file content to ensure the generated file aligns with the user requirement:
$<$similar\_case\_reference$>$\{similar\_file\_text\}$<$/similar\_case\_reference$>$. 
Similar case reference is always correct. If you find the user requirement is very consistent with the similar case reference, you should use the similar case reference as the template to generate the file.
Just modify the necessary parts to make the file complete and functional.
Please ensure that the generated file is complete, functional, and logically sound.
Additionally, apply your domain expertise to verify that all numerical values are consistent with the user requirements, maintaining accuracy and coherence.

You should ensure that the new file is consistent with the previous files. Such as boundary conditions, mesh settings, etc.
\end{tcolorbox}

\subsubsection{Command Generation Prompts}

\begin{tcolorbox}[colback=blue!5!white,colframe=blue!75!black,title=Command Generation System Prompt]
You are an expert in OpenFOAM. The user will provide a list of available commands. 
Your task is to generate only the necessary OpenFOAM commands required to create an Allrun script for the given user case, based on the provided directory structure. 
Return only the list of commands -- no explanations, comments, or additional text.
\end{tcolorbox}

\begin{tcolorbox}[colback=blue!5!white,colframe=blue!75!black,title=Command Generation User Prompt]
Available OpenFOAM commands for the Allrun script: \{commands\}\\
Case directory structure: \{state.dir\_structure\}\\
User case information: \{state.case\_info\}\\
Reference Allrun scripts from similar cases: \{state.allrun\_reference\}\\
Generate only the required OpenFOAM command list---no extra text.
\end{tcolorbox}
The commands passed in the user prompt to this agent form a pre-curated list of OpenFOAM commands, saved within \fa. The user need not provide any command within the input. 
\subsubsection{Allrun Script Generation Prompts}

\begin{tcolorbox}[colback=blue!5!white,colframe=blue!75!black,title=Allrun Script Generation System Prompt]
You are an expert in OpenFOAM. \\Generate an Allrun script based on the provided details.\\
Available commands with descriptions: \{commands\_help\}\\
Reference Allrun scripts from similar cases: \{state.allrun\_reference\}
\end{tcolorbox}

\begin{tcolorbox}[colback=blue!5!white,colframe=blue!75!black,title=Allrun Script Generation User Prompt]
User requirement: \{state.user\_requirement\}\\
Case directory structure: \{state.dir\_structure\}\\
User case information: \{state.case\_info\}\\
All run scripts for these similar cases are for reference only and may not be correct, as you might be a different case solver or have a different directory structure. 
You need to rely on your OpenFOAM and physics knowledge to discern this, and pay more attention to user requirements, 
as your ultimate goal is to fulfill the user's requirements and generate an allrun script that meets those requirements.
Generate the Allrun script strictly based on the above information. Do not include explanations, comments, or additional text. Put the code in \texttt{\textasciigrave \textasciigrave \textasciigrave} tags.
\end{tcolorbox}

\subsection{Reviewer Agent Prompts}

The Reviewer Agent analyzes simulation errors and proposes corrections to resolve issues.

\subsubsection{Error Analysis Prompts}

\begin{tcolorbox}[colback=blue!5!white,colframe=blue!75!black,title=Error Analysis System Prompt]
You are an expert in OpenFOAM simulation and numerical modeling. 
Your task is to review the provided error logs and diagnose the underlying issues. 
You will be provided with a similar case reference, which is a list of similar cases that are ordered by similarity from high to low. You can use this reference to help you understand the user requirement and the error.
When an error indicates that a specific keyword is undefined (for example, `div(phi,(p$\vert$rho)) is undefined'), your response must propose a solution that simply defines that exact keyword as shown in the error log. 
Do not reinterpret or modify the keyword (e.g., do not treat \texttt{div(phi,(p|rho))} as meaning ``or''); instead, assume it is meant to be taken literally.
Propose ideas on how to resolve the errors, but do not modify any files directly. 
Please do not propose solutions that require modifying any parameters declared in the user requirement, try other approaches instead. Do not ask the user any questions.
The user will supply all relevant foam files along with the error logs, and within the logs, you will find both the error content and the corresponding error command indicated by the log file name.
\end{tcolorbox}

\begin{tcolorbox}[colback=blue!5!white,colframe=blue!75!black,title=Error Analysis User Prompt (Initial Error)]
$<$similar\_case\_reference$>$\{state.tutorial\_reference\}$<$/similar\_case\_reference$>$\\
$<$foamfiles$>$\{str(state.foamfiles)\}$<$/foamfiles$>$\\
$<$error\_logs$>$\{state.error\_logs\}
$<$/error\_logs$>$\\
$<$user\_requirement$>$\{state.user\_requirement\}$<$/user\_requirement$>$\\
Please review the error logs and provide guidance on how to resolve the reported errors. Make sure your suggestions adhere to user requirements and do not contradict it.
\end{tcolorbox}

\begin{tcolorbox}[colback=blue!5!white,colframe=blue!75!black,title=Error Analysis User Prompt (Subsequent Errors)]
$<$similar\_case\_reference$>$\{state.tutorial\_reference\}$<$/similar\_case\_reference$>$\\
$<$foamfiles$>$\{str(state.foamfiles)\}$<$/foamfiles$>$\\
$<$current\_error\_logs$>$\{state.error\_logs\}
$<$/current\_error\_logs$>$\\
$<$history$>$
\{chr(10).join(state.history\_text)\}
$<$/history$>$\\
$<$user\_requirement$>$\{state.user\_requirement\}$<$/user\_requirement$>$\\
I have modified the files according to your previous suggestions. If the error persists, please provide further guidance. Make sure your suggestions adhere to user requirements and do not contradict them. Also, please consider the previous attempts and try a different approach.
\end{tcolorbox}

\subsubsection{File Correction Prompts}

\begin{tcolorbox}[colback=blue!5!white,colframe=blue!75!black,title=File Correction System Prompt]
You are an expert in OpenFOAM simulation and numerical modeling. 
Your task is to modify and rewrite the necessary OpenFOAM files to fix the reported error. 
Please do not propose solutions that require modifying any parameters declared in the user requirement; try other approaches instead.
The user will provide the error content, error command, reviewer's suggestions, and all relevant foam files. 
Only return files that require rewriting, modification, or addition; do not include files that remain unchanged. 
Return the complete, corrected file contents in the following JSON format: \\
\{\\
list of foamfile: [\{file\_name: `file\_name',\\
folder\_name: `folder\_name',\\
content: `content'\}]. \\
\} \\
Ensure your response includes only the modified file content with no extra text, as it will be parsed using Pydantic.
\end{tcolorbox}

\begin{tcolorbox}[colback=blue!5!white,colframe=blue!75!black,title=File Correction User Prompt]
$<$foamfiles$>$\{str(state.foamfiles)\}$<$/foamfiles$>$\\
$<$error\_logs$>$\{state.error\_logs\}$<$/error\_logs$>$\\$<$reviewer\_analysis$>$\{review\_content\}$<$/reviewer\_analysis$>$\\
$<$user\_requirement$>$\{state.user\_requirement\}$<$/user\_requirement$>$\\

Please update the relevant OpenFOAM files to resolve the reported errors, ensuring that all modifications strictly adhere to the specified formats. Ensure all modifications adhere to user requirement.
\end{tcolorbox}

\subsection{History Tracking Format}

The system tracks modification history using a structured format for each iteration attempt:
\begin{tcolorbox}[colback=blue!5!white,colframe=blue!75!black,title=History Tracking Format]
$<$Attempt \{attempt\_number\}$>$\\
$<$Error\_Logs$>$
\{state.error\_logs\}
$<$/Error\_Logs$>$\\
$<$Review\_Analysis$>$
\{review\_content\}
$<$/Review\_Analysis$>$\\
$<$/Attempt$>$
\end{tcolorbox}

\subsection{Example User Requirements}

Below is an example of a user requirement used to test the \fa~system:

\begin{tcolorbox}[colback=blue!5!white,colframe=blue!75!black,title=Example User Requirement]
Perform a 3D Benard Cell simulation using OpenFOAM. 
Run a two-dimensional Rayleigh Benard convection cell simulation in a rectangular cavity that is 9 units wide and 1 unit tall, extruded a single cell in the spanwise direction with the front and back patches treated as empty to enforce 2D. Use 90 cells in x direction, 10 cells in y and 1 in z. Use the buoyantFoam solver with gravity acting downward in the negative y direction. The reference state should be air-like with density 1.0 kg/m$^3$, reference temperature 300 K, thermal expansion coefficient 3.3e-3 1/K, kinematic viscosity 1.5e-5 $m^2/s$, and thermal diffusivity 2.1e-5 $m^2/s$ (Pr = 0.7). Temperature of bottom wall is fixed at 301 K and the top wall at 300 K, while the vertical side walls are adiabatic. Apply no-slip velocity conditions on all solid walls, initialize the flow at rest with a uniform temperature field of 300 K. Use a timestep of 1s, simulate up to 1000 s of physical time, and write results every 50 steps. Use k-epsilon turbulence model.

\end{tcolorbox}

\pagebreak
\section{User Prompts In Case Studies}
\label{app:b}

\phantomsection
\subsection{Multi Element Airfoil}
\begin{tcolorbox}[colback=blue!5!white,colframe=blue!75!black,title=Example User Requirement]
Perform a 2D incompressible flow over a multi element airfoil setup. The mesh is provided as a \texttt{.msh} file. The \texttt{.msh} file contains 4 boundaries named ``inlet'',
``outlet'', ``walls'', ``airfoil'' and ``frontAndBack''. The ``inlet'' and ``outlet'' are of type freestream with the freestream velocity being 9 m/s. 
 The ``walls'' and ``airfoil'' have a no-slip boundary condition
  (velocity equal to zero at the wall). The ``frontAndBack''  faces are designated as
  `empty'. The simulation runs from time 0 to 1000 with a time step of 1.0 units,
  and results are output every 1 time step. The viscosity (nu) is set as constant
  with a value of $1.5\times10^{-5}$ $\mathrm{m^2/s}$. Use simpleFoam solver. Use SpalartAllmaras turbulence model. Further visualize the magnitude of velocity along the Z plane.
\end{tcolorbox}

\phantomsection
\subsection{Tandem Wing}
\begin{tcolorbox}[colback=blue!5!white,colframe=blue!75!black,title=Example User Requirement]
Perform a 3D incompressible flow over a tandem wing configuration. The mesh is provided as a \texttt{.msh} file. The msh file contains 4 boundaries named ``inlet'',
``outlet'', ``walls'', ``airfoil'' and ``frontAndBack''. The ``inlet'' and ``outlet'' are of type freestream with the freestream velocity being 9 m/s. 
 The ``walls'' and ``airfoil'' have a no-slip boundary condition (velocity equal to zero at the wall). The ``frontAndBack''  faces are also of type wall. 
 The simulation runs from time 0 to 1000 with a time step of 1.0 units,
  and results are output every 1 time step. The viscosity (`nu') is set as constant
  with a value of $1.5\times10^{-5}$ $\mathrm{m^2/s}$. Use simpleFoam solver. Use SpalartAllmaras turbulence model. Further visualize the magnitude of velocity along the mid Z section at the final time.
\end{tcolorbox}

\phantomsection
\subsection{Flow Over Cylinder}
\begin{tcolorbox}[colback=blue!5!white,colframe=blue!75!black,title=Example User Requirement]
Simulate incompressible flow over a circular cylinder. 
Use Gmsh to create the computational mesh.
The computational domain extends from -2.5 to 2.5 in the x-direction, -1 to 1 in the y-direction, and 0 to 0.2 in the z-direction. 
The cylinder is positioned at (-1, 0) with a radius of 0.1 units. 
Use a structured mesh with approximately 20x10 cells in the x-y plane and 1 cell in the z-direction. 
The inlet boundary named ``inlet'' (left boundary at x = -2.5) has a uniform velocity of 1 m/s in the positive x-direction. 
The right boundary at x=+2.5 is the outlet named ``outlet''. 
The top and bottom walls named ``topWall'' and ``bottomWall'' respectively (y = +1 and y=-1) use slip boundary conditions. 
The cylinder surface named ``cylinder'' uses a no-slip boundary condition (velocity equal to zero at the wall). 
The front and back faces named ``frontAndBack'' are located at z = 0 and z = 0.2 respectively, and are designated as `empty' for 2D simulation.  
Use base mesh size of 0.5 on cylinder and size of 1.0 elsewhere.
The simulation runs from time 0 to 2 seconds with a time step of 0.001 units, and results are output every 100 time steps. 
The kinematic viscosity (nu) is set as constant with a value of $1\times10^{-5}$ $\mathrm{m^2/s}$. Use pisoFoam solver for incompressible flow.
Visualize the magnitude of velocity ('U') along the x-y plane. 
\end{tcolorbox}

\phantomsection
\subsection{Flow Over Two Square Obstacles}
\begin{tcolorbox}[colback=blue!5!white,colframe=blue!75!black,title=Example User Requirement]
Simulate incompressible flow over two square obstacles. 
Use Gmsh to create the computational mesh.
The computational domain spans 0 to 5 in x direction and 0 to 2.5 in y direction and 0 to 0.1 in the z direction. 
One of the square obstacle is of size 0.25 unit x 0.25 unit x 0.1 unit centered at 1.5 x 1.25 x 0.0 and the other square obstacle is of size 0.25 unit x 0.25 unit x 0.1 unit centered at 3.5 x 1.25 x 0.0. 
Use one cell in z direction making the geometry effectively 2D. 
Use a structured mesh with approximately 50x25 cells in the x-y plane and 1 cell in the z-direction. 
The inlet boundary named ``inlet'' (left boundary at x = 0) has a uniform velocity of 1 m/s in the positive x-direction. 
The right boundary at x = 5 is the outlet named ``outlet''. 
The top and bottom walls named ``topWall'' and ``bottomWall'' respectively (y = 2.5 and y = 0) use slip boundary conditions. 
The square obstacle surfaces named ``square1'' and ``square2'' use no-slip boundary conditions (velocity equal to zero at the walls). 
The front and back faces named ``frontAndBack'' are located at z = 0 and z = 0.1 respectively, and are designated as `empty' for 2D simulation.  
Use base mesh size of 0.5 on squares and size of 1.0 elsewhere.
The simulation runs from time 0 to 10 seconds with a time step of 0.001 units, and results are output every 100 time steps. 
The kinematic viscosity (nu) is set as constant with a value of $1\times10^{-5}$ $\mathrm{m^2/s}$. Use pisoFoam solver for incompressible flow.
Visualize the magnitude of velocity (`U') along the x-y plane. 
\end{tcolorbox}

\phantomsection
\subsection{3D cavity HPC Case}
\begin{tcolorbox}[colback=blue!5!white,colframe=blue!75!black,title=Example User Requirement]
Do an incompressible 3D lid driven cavity flow using icoFoam solver. 
The cavity is a cube of dimension [0, 0.1] $\times$ [0, 0.1] $\times$ [0,0.1]. 
Use simple grading with 100 $\times$100$\times$100 in $x$, $y$ and $z$ direction. 
The top wall (`movingWall') moves in the x-direction with a uniform velocity of 1 m/s. 
The `fixedWalls' have a no-slip boundary condition (velocity equal to zero at the wall). 
The simulation runs from time 0 to 0.015 with a time step of 0.001 units, and results are output every 10 time steps. 
The viscosity (nu) is set as constant with a value of 0.01 $\mathrm{m^2/s}$.
Perform a HPC run for this case in perlmutter cluster. My account is ACCOUNTNAME. Do a parallel run for this case by splitting it into 32 subdomains. 
\end{tcolorbox}

\pagebreak
\section{Slurm Script Generated by the Agent}      
\label{app:c}
        
\begin{tcolorbox}[colback=blue!5!white,colframe=blue!75!black,title=Slurm Script Generated by the Agent]

\#!/bin/bash\\
\#SBATCH -A xxxx\\
\#SBATCH -C cpu\\
\#SBATCH -q regular\\
\#SBATCH -N 1\\
\#SBATCH -n 32\\
\#SBATCH --ntasks-per-node=32\\
\#SBATCH -t 02:00:00\\
\#SBATCH --mem=128GB\\
\#SBATCH -J Sim\\
\#SBATCH -o \%j.out\\
\#SBATCH -e \%j.err\\

\# Initialize error handling\\
set -e\\

\# Change to case directory\\
cd ./runshpcnode $\mid\mid$ exit 1\\

\# Create log directory if it doesn't exist\\
mkdir -p logs\\

\# Run the simulation\\
echo ``Starting OpenFOAM simulation at \$(date)''\\
./Allrun -parallel\\
if [ \$? -eq 0 ]; then\\
    echo ``Simulation completed successfully at \$(date)''\\
    exit 0\\
else\\
    echo ``Simulation failed at \$(date)''\\
    exit 1\\
fi
\end{tcolorbox}

\pagebreak
\section{MCP Functions}
\label{app:d}
\begin{figure*}[htbp]
    \centering
    \includegraphics[width=0.9\textwidth]{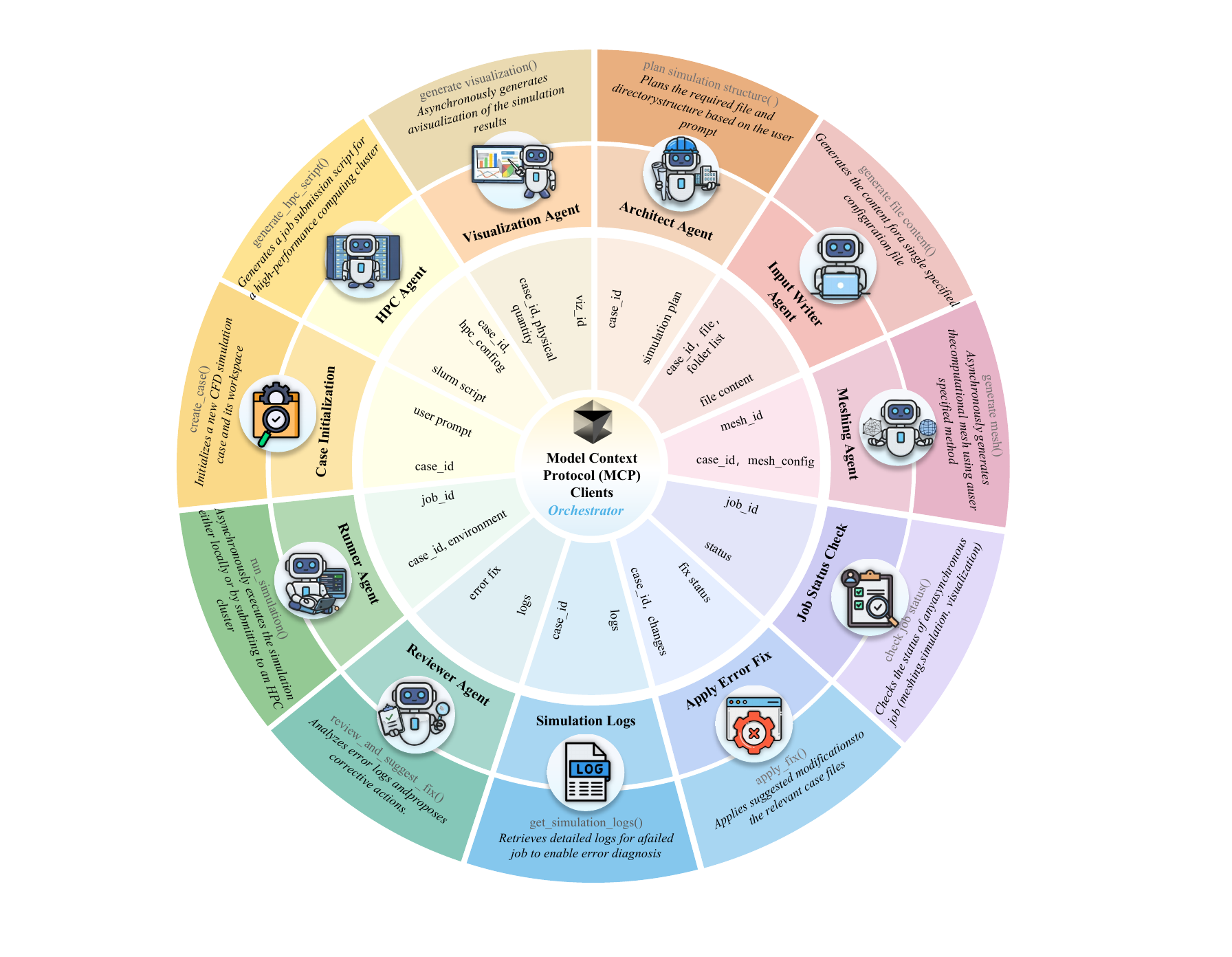}
    \caption{Design diagram for the MCP version of the \fa. Functionality of \fa is broken down to basic functions and exposed to an MCP orchestrator. The orchestrator can then decide which function to be called at anytime, given the user request.}
    \label{fig:mcp}
\end{figure*}

The purpose of each function in the MCP compliant architecture of \fa is shown in \Cref{tab:mcp_functions}; the overall MCP design, in which \fa's functionality is decomposed into basic functions and exposed to an external orchestrator, is shown in \Cref{fig:mcp}.

\begin{table}[H]
\small
\centering
\caption{The core functions of the \fa~Model Context Protocol (MCP). Each function represents a decoupled capability within the CFD workflow, featuring strongly-typed inputs and outputs to ensure reliable interaction with orchestrating agents.}
\label{tab:mcp_functions}
\resizebox{\textwidth}{!}{%
\begin{tabular}{>{\raggedright\arraybackslash}p{0.3\textwidth} >{\raggedright\arraybackslash}p{0.29\textwidth} >{\raggedright\arraybackslash}p{0.15\textwidth} >{\raggedright\arraybackslash}p{0.20\textwidth}}
\toprule
\textbf{Function Name} & \textbf{Description} & \textbf{Input Schema} & \textbf{Output Schema} \\
\midrule
\texttt{create\_case} & Initializes a new CFD simulation case and its workspace. & \texttt{\{user\_prompt: str\}} & \texttt{\{case\_id: str\}} \\
\addlinespace
\texttt{plan\_simulation\_structure} & (Architect Agent) Plans the required file and directory structure based on the user prompt. & \texttt{\{case\_id: str\}} & \texttt{\{plan: List[\{file, folder\}]\}} \\
\addlinespace
\texttt{generate\_file\_content} & (Input Writer Agent) Generates the content for a single specified configuration file. & \texttt{\{case\_id, file, folder\}} & \texttt{\{content: str\}} \\
\addlinespace
\texttt{generate\_mesh} & (Meshing Agent) Asynchronously generates the computational mesh using a specified method. & \texttt{\{case\_id, mesh\_config: Dict\}} & \texttt{\{job\_id: str\}} \\
\addlinespace
\texttt{generate\_hpc\_script} & (HPC Agent) Generates a job submission script (e.g., Slurm) for a high-performance computing cluster. & \texttt{\{case\_id, hpc\_config: Dict\}} & \texttt{\{script\_content: str\}} \\
\addlinespace
\texttt{run\_simulation} & (Runner Agent) Asynchronously executes the simulation either locally or by submitting to an HPC cluster. & \texttt{\{case\_id, environment: str\}} & \texttt{\{job\_id: str\}} \\
\addlinespace
\texttt{check\_job\_status} & Checks the status of any asynchronous job (meshing, simulation, visualization). & \texttt{\{job\_id: str\}} & \texttt{\{status: Dict\}} \\
\addlinespace
\texttt{get\_simulation\_logs} & Retrieves detailed logs for a failed job to enable error diagnosis. & \texttt{\{case\_id, job\_id\}} & \texttt{\{logs: Dict\}} \\
\addlinespace
\texttt{review\_and\_suggest\_fix} & (Reviewer Agent) Analyzes error logs and proposes corrective actions. & \texttt{\{case\_id, logs\}} & \texttt{\{suggestions: Dict\}} \\
\addlinespace
\texttt{apply\_fix} & Applies suggested modifications to the relevant case files. & \texttt{\{case\_id, modifications: List\}} & \texttt{\{status: str\}} \\
\addlinespace
\texttt{generate\_visualization} & (Visualization Agent) Asynchronously generates a visualization of the simulation results. & \texttt{\{case\_id, quantity, ...\}} & \texttt{\{job\_id: str\}} \\
\bottomrule
\end{tabular}%
}
\end{table}

\paragraph{Protocol-level characteristics.} Two protocol-boundary concerns are documented here for completeness. \emph{Latency.} Per-call round-trip overhead is small in practice (single-digit milliseconds for local STDIO transports and tens of milliseconds for HTTP transports) and is dominated end-to-end by OpenFOAM solver wall-clock per Reviewer iteration (\Cref{sec:cost}). \emph{Protocol-layer failure modes.} Four error classes arise at the protocol boundary: schema-validation errors on malformed argument payloads, call timeouts when a long-running solver invocation exceeds the orchestrator-side timeout, solver-crash propagation when an OpenFOAM signal exit is surfaced through a tool-call return, and orchestrator session drops during long Reviewer loops. Each is returned as a structured error payload via the MCP error contract, preserving \fa's standalone error semantics across the boundary.

\section{Supplementary Experiments on Additional Backbones}
\label{app:suppl}

The two studies reported in this appendix were run on backbones other than the Claude~3.5~Sonnet model used for the headline FoamBench evaluation---a prompt-sensitivity study on GPT-5.3-codex and a per-mesh-type reliability breakdown on Claude Sonnet~4.6---and are presented separately, as relative within-study comparisons.

\subsection{Prompt Sensitivity}
\label{sec:promptsens}
Because \fa~is driven by natural language, we test robustness to prompt phrasing. For each of the $27$ FoamBench datasets (11 Basic-tier scenarios and 16 Advanced-tier cases) we generated three LLM paraphrases of the prompt under a syntactic-drift guard that preserves all numerics, single-quoted boundary-condition names, and OpenFOAM solver names; the median embedding-space faithfulness to the original prompt is $0.959$ (cosine similarity). \Cref{tab:promptsens} summarizes the stability: the bucketed $M_{\mathrm{NMSE}}$ agrees across all three paraphrases for $18/27$ datasets and $M_{\mathrm{exec}}$ for $22/27$, the Advanced-tier bucketed $M_{\mathrm{NMSE}}$ mean shows zero variation, and the Basic-tier mean varies by at most $0.09$. The $5/27$ datasets with $M_{\mathrm{exec}}$ disagreement concentrate in known-hard cases (e.g.\ \texttt{forwardStep} and \texttt{wedge} on Basic, \texttt{Rectangular\_Obstacle} variants on Advanced), indicating case difficulty rather than phrasing sensitivity. We read this as evidence of semantic-equivalence robustness; lexical-diversity robustness across model families and manual rewrites is a complementary evaluation.

\begin{table}[tbp]
\centering
\caption{Prompt-sensitivity robustness on the GPT-5.3-codex backbone: three LLM paraphrases per dataset over the $27$ FoamBench datasets.}
\label{tab:promptsens}
\begin{tabular}{@{}p{0.66\columnwidth}r@{}}
\toprule
Robustness check & Result \\
\midrule
$M_{\mathrm{NMSE}}$ bucket agreement across all 3 paraphrases & 18/27 \\
$M_{\mathrm{exec}}$ agreement across all 3 paraphrases & 22/27 \\
Advanced-tier bucketed $M_{\mathrm{NMSE}}$ mean variation & 0 \\
Basic-tier bucketed $M_{\mathrm{NMSE}}$ mean variation & $\le 0.09$ \\
\bottomrule
\end{tabular}
\end{table}

\subsection{Per-Mesh-Type Reliability}
To characterize mesh-type reliability we instrument a FoamBench-Advanced run ($n{=}16$) with per-invocation mesh-type metadata (\Cref{tab:meshtype}); FoamBench-Basic is uniformly \texttt{blockMesh}, so the Basic numbers already report blockMesh performance. Two observations follow. The \texttt{snappyHexMesh} Advanced cases reach $M_{\mathrm{exec}}=0.80$ but cluster in fidelity bucket~0: generating a topologically valid hex-dominant mesh is not the bottleneck on this path---the residual is solver-side physical accuracy. The Advanced \texttt{blockMesh} cases are bimodal, with one complete-success case and the remainder split between divergent and execution-failure outcomes. In both pathways the limiting factor on the Advanced tier is solver fidelity and parameter setup, not mesh generation. The absolute values in \Cref{tab:meshtype} are reported for the within-table mesh-type comparison.

\begin{table}[tbp]
\centering
\caption{Per-mesh-type success on FoamBench-Advanced ($n{=}16$), Claude Sonnet~4.6 backbone. ``buckets'' are the case counts in fidelity buckets $1/0.5/0$.}
\label{tab:meshtype}
\begin{tabular}{lcccc}
\toprule
Mesh kind & Cases & $M_{\mathrm{exec}}$ & $M_{\mathrm{NMSE}}$ & buckets ($1/0.5/0$) \\
\midrule
\texttt{blockMesh}     & 11 & 0.727 & 0.091 & $1/0/10$ \\
\texttt{snappyHexMesh} &  5 & 0.800 & 0.000 & $0/0/5$ \\
\bottomrule
\end{tabular}
\end{table}

\end{document}